%% file: emnlp2023.tex
\pdfoutput=1

\documentclass[11pt]{article}
\usepackage{tabularx}

\usepackage[final]{EMNLP2023}

\usepackage{times}
\usepackage{latexsym}

\usepackage[T1]{fontenc}

\usepackage[utf8]{inputenc}

\usepackage{microtype}

\usepackage{inconsolata}

\usepackage{todonotes}
\usepackage{tcolorbox}
\usepackage{tabularx}
\usepackage{array} 
\usepackage{makecell} 
\usepackage{graphicx}
\usepackage{subcaption}
\usepackage{booktabs} 
\usepackage{xcolor}
\usepackage{lipsum}
\usepackage{geometry}
\usepackage{amsmath}
\usepackage{multirow}
\usepackage{caption}
\usepackage[dvipsnames]{xcolor}
\usepackage{tabularx}
\usepackage{ragged2e}
\usepackage{siunitx} 
\usepackage{amssymb}
\usepackage{tablefootnote}

\tcbuselibrary{skins} 

\title{BLEnD-Vis: Benchmarking Multimodal Cultural Understanding in \\ Vision Language Models}

\author{
  Bryan Chen Zhengyu Tan$^{1,2}$\thanks{\hspace{2.5mm}Equal contribution.} \and
  Weihua Zheng$^{1,2}$\footnotemark[1] \\  
 \textbf{ Zhengyuan Liu$^{2}$ 
  Nancy F. Chen$^{2}$ 
  Hwaran Lee$^{3}$ 
  Kenny Tsu Wei Choo$^{1}$ 
  Roy Ka-Wei Lee$^{1}$} \\ 
  \begin{tabular}{c} 
    $^{1}$Singapore University of Technology and Design (SUTD) \\
    $^{2}$Institute for Infocomm Research (I2R), A*STAR, Singapore \\
    $^{3}$Sogang University \\
  \end{tabular}
}

\begin{document}
\maketitle

\begin{abstract}

As vision-language models (VLMs) are deployed globally, their ability to understand culturally situated knowledge becomes essential. Yet, existing evaluations largely assess static recall or isolated visual grounding, leaving unanswered whether VLMs possess robust and transferable cultural understanding. We introduce \textbf{\texttt{BLEnD-Vis}}, a multimodal, multicultural benchmark designed to evaluate the \textit{robustness} of everyday cultural knowledge in VLMs across linguistic rephrasings and visual modalities. Building on the BLEnD dataset, \textbf{\texttt{BLEnD-Vis}} constructs 313 culturally grounded question templates spanning 16 regions and generates three aligned multiple-choice formats: (i) a text-only baseline querying from Region $\rightarrow$ Entity, (ii) an inverted text-only variant (Entity $\rightarrow$ Region), and (iii) a VQA-style version of (ii) with generated images. The resulting benchmark comprises 4,916 images and over 21,000 multiple-choice questions (MCQ) instances, validated through human annotation. \textbf{\texttt{BLEnD-Vis}} reveals significant fragility in current VLM cultural knowledge; models exhibit performance drops under linguistic rephrasing. While visual cues often aid performance, low cross-modal consistency highlights the challenges of robustly integrating textual and visual understanding, particularly in lower-resource regions. \textbf{\texttt{BLEnD-Vis}} thus provides a crucial testbed for systematically analysing cultural robustness and multimodal grounding, exposing limitations and guiding the development of more culturally competent VLMs. Code is available at https://github.com/Social-AI-Studio/BLEnD-Vis.

\end{abstract}

\input{sections/1_intro}

\input{sections/2_related_works}

\input{sections/3_data}
\input{sections/4_results}
\input{sections/5_discussion}
\input{sections/6_conclusion}

\input{sections/7_limitations}

\input{sections/8_ethical_statement}

\input{sections/9_acknowledgement}

\bibliography{anthology,emnlp2023}
\bibliographystyle{acl_natbib}

\appendix
\label{sec:appendix}
\input{sections/10_appendix}

\end{document}

%% file: sections/1_intro.tex
\section{Introduction}
\label{sec:intro}
\begin{figure*}[!t]
    \centering
    \includegraphics[width=\linewidth]{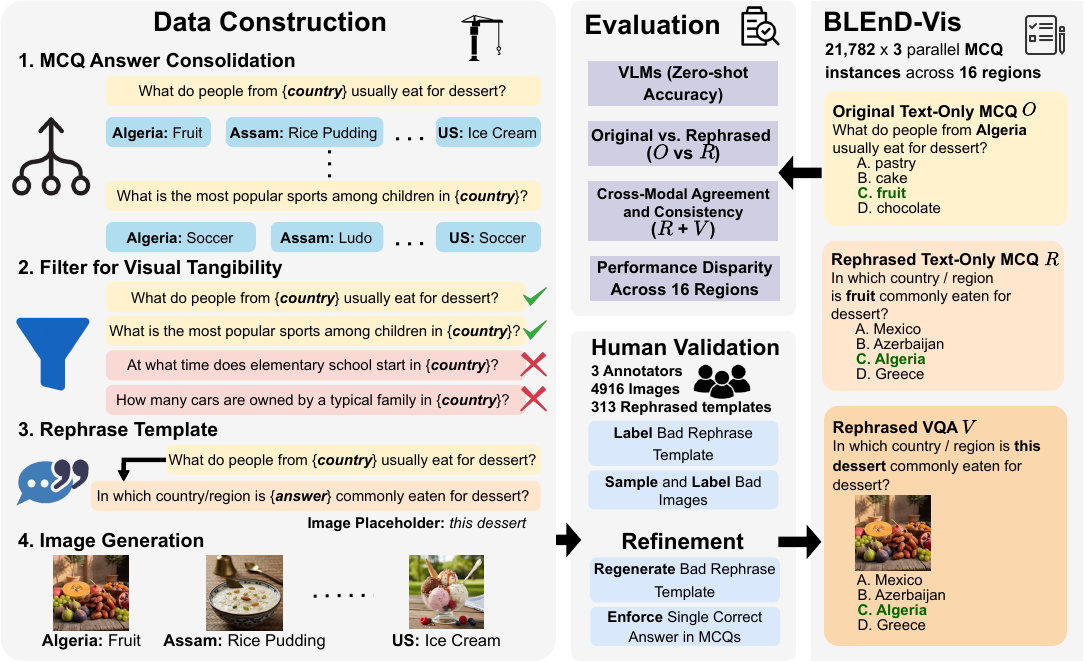} 
    \caption{Overview of the \textbf{\texttt{BLEnD-Vis}} benchmark construction and evaluation framework. The process involves: (1) \textbf{Data Construction} via tangibility filtering, question rephrasing, and image generation based on BLEnD; (2) \textbf{Human Validation} of generated assets; (3) Creation of the final \textbf{\texttt{BLEnD-Vis} Dataset} comprising three parallel MCQ formats (Original Text, Rephrased Text, VQA) across 16 regions; and (4) \textbf{Evaluation} assessing VLM zero-shot accuracy, robustness to rephrasing, cross-modal consistency, and regional performance variations.}
    \label{fig:overview}
\end{figure*}

As large language models (LLMs) and vision-language models (VLMs) become increasingly embedded in global applications, their capacity to comprehend and respond to diverse cultural contexts is gaining critical importance \citep{pawarSurveyCulturalAwareness2024, adilazuardaMeasuringModelingCulture2024, liSurveyStateArt2025}. While these models exhibit impressive general capabilities, they often falter in understanding everyday cultural practices—such as local foods, leisure activities, and family customs—particularly for communities that are underrepresented in mainstream training corpora \citep{NEURIPS2024_8eb88844, alkhamissiInvestigatingCulturalAlignment2024}. This poses real-world risks, as cultural insensitivity can undermine user trust, marginalise minority populations, and perpetuate global inequities in AI deployment \citep{qiuEvaluatingCulturalSocial2025, kannenAestheticsCulturalCompetence2024}.

Existing benchmarks that aim to evaluate cultural knowledge typically assess recall via direct textual prompts \citep{NEURIPS2024_9a16935b, wangCDEvalBenchmarkMeasuring2024} or focus on narrow multimodal contexts \citep{urailertprasertSEAVQASoutheastAsian2024, nayakBenchmarkingVisionLanguage2024, satarSeeingCultureBenchmark2025, winataWorldCuisinesMassivescaleBenchmark2025}. Yet, two essential questions remain underexplored: (1) How robust are these models to linguistic rephrasings of culturally grounded queries? (2) Can they consistently ground cultural knowledge in visual representations? These questions are important for distinguishing deep conceptual understanding from superficial or brittle associations that may degrade under linguistic or visual variation \citep{zhangBAVIBenchEvaluatingRobustness2025, leeVHELMHolisticEvaluation2024}.

To bridge this gap, we propose \textbf{\texttt{BLEnD-Vis}}, a multimodal, multicultural benchmark designed to evaluate the robustness and groundedness of everyday cultural knowledge in VLMs. \textbf{\texttt{BLEnD-Vis}} builds upon the BLEnD dataset \citep{NEURIPS2024_8eb88844} by selecting a curated set of tangible, culturally situated concepts across 16 diverse regions and constructing three parallel evaluation formats: \textit{Original MCQ}, \textit{Rephrased MCQ}, and \textit{VQA-Style MCQ}. These formats enable controlled comparisons that isolate the effects of linguistic rephrasing and modality shifts on model performance. While this study focuses on English to leverage the strength of state-of-the-art models and ensure comparability across formats, \textbf{\texttt{BLEnD-Vis}} sets the stage for future multilingual expansion. 

Our evaluation of twelve VLMs on \textbf{\texttt{BLEnD-Vis}} reveals that: (i) model performance does not strongly correlate with parameter count; (ii) linguistic rephrasing often degrades performance, indicating brittle knowledge representations; (iii) visual cues can significantly aid understanding, with VQA performance often surpassing text-only rephrased queries; and (iv) robust cross-modal consistency (correctness on both $R$ and $V$ formats for the same fact) is particularly challenging, with a mean joint correctness of only 42.19\%. Furthermore, preliminary cross-modal fine-tuning experiments indicate that training on textual data generally improves VQA performance (mean +16.15\%), while the transfer from VQA training to textual performance is less apparent (mean +3.98\%). Figure \ref{fig:overview} shows the overview of the BLEnD-Vis benchmark construction and evaluation framework. We summarise our contributions as follows:

\begin{itemize}
    \item We present \textbf{\texttt{BLEnD-Vis}}, a benchmark to evaluate the robustness of everyday cultural knowledge in VLMs across linguistic and visual modalities.
    \item We develop a systematic pipeline involving tangibility filtering, question rephrasing, image generation, and human validation.
    \item We release a dataset of 313 validated templates, 4916 culturally grounded images, and 21,782 MCQs in three aligned formats.
    \item We provide a comparative evaluation of twelve VLMs, revealing gaps in cultural robustness and cross-modal generalisation.
\end{itemize}

\textbf{\texttt{BLEnD-Vis}} provides a rigorous framework to assess whether state-of-the-art VLMs encode not only cultural facts, but robust and transferable cultural understanding in both language and vision.



%% file: sections/2_related_works.tex
\section{Related Works}
\label{sec:related_work}

\subsection{Cultural Knowledge Benchmarks}
Recent benchmarks assess cultural knowledge in LLMs through textual recall or alignment. BLEnD \citep{NEURIPS2024_8eb88844} captures everyday knowledge across regions and languages; CultureLLM \citep{NEURIPS2024_9a16935b}, CulturePark \citep{NEURIPS2024_77f089cd}, and CDEval \citep{wangCDEvalBenchmarkMeasuring2024} measure alignment with cultural dimensions. SeaEval \citep{wangSeaEvalMultilingualFoundation2024} and MAPS \citep{liuAreMultilingualLLMs2024} extend this to multilingual or proverb-based reasoning. Palm \cite{alwajihPalmCulturallyInclusive2025} provides a culturally inclusive dataset but focuses on Arabic LLMs and textual modalities. RAVENEA \cite{liRAVENEABenchmarkMultimodal2025} introduces a multimodal benchmark but prioritises Retrieval-Augmented Generation (RAG). While these benchmarks focus on cultural knowledge, they do not evaluate robustness to rephrasing or modality shifts. In contrast, \textbf{\texttt{BLEnD-Vis}} probes the robustness of cultural understanding through controlled textual rephrasing and image grounding, offering deeper diagnostic insights.

\subsection{Cultural Bias and Alignment}
Cultural bias in LLMs and VLMs is an ongoing concern, with studies showing that models often reflect Western-centric norms \citep{taoCulturalBiasCultural2024}. Alignment methods have sought to mitigate this, usually using human-annotated survey responses as ground truth \citep{alkhamissiInvestigatingCulturalAlignment2024}, or exposing covert harms through social scenario simulation \citep{dammuTheyAreUncultured2024, tanUnmaskingImplicitBias2025}. Other work explores the causal assumptions embedded in culturally-influenced data \citep{bauerSocialCommonsenseExplanation2023} and the fragility of model outputs when subjected to adversarial prompts \citep{zhangBAVIBenchEvaluatingRobustness2025}. Although these approaches surface latent biases, \textbf{\texttt{BLEnD-Vis}} complements them by testing representational stability across phrasings and visual formats, exposing brittle associations and performance disparities across diverse cultures.

\subsection{Multimodal Cultural Evaluation}
Multimodal benchmarks like SEA-VQA \citep{urailertprasertSEAVQASoutheastAsian2024} and CulturalVQA \citep{nayakBenchmarkingVisionLanguage2024} test VLMs on culturally situated images, often exposing regional performance gaps. ALM-Bench \citep{vayaniAllLanguagesMatter2025} broadens this across 100 languages. Others evaluate cultural competence in generative models \citep{kannenAestheticsCulturalCompetence2024, dincaOpenBiasOpensetBias2024} or few-shot adaptation \citep{nikandrouCROPEEvaluatingIncontext2025, kimWHENTOMEATS2025}. \textbf{\texttt{BLEnD-Vis}} is distinct in aligning textual and visual formats for the same cultural fact, enabling fine-grained comparisons and disentangling the effects of linguistic versus perceptual variation.

%% file: sections/3_data.tex
\section{\textbf{\texttt{BLEnD-Vis}} Dataset Construction}
\label{sec:dataset_curation}

The \textbf{\texttt{BLEnD-Vis}} benchmark is constructed via a multi-stage pipeline that transforms the textual, everyday cultural knowledge from BLEnD \citep{NEURIPS2024_8eb88844} into parallel evaluation sets suitable for probing textual and multimodal robustness.

\subsection{Base Dataset and Scope}
\label{sec:base_dataset_scope}
We utilise the 500 English Short Answer Question (SAQ) templates derived from the original BLEnD study as our starting point. The ground truth answers for these templates are sourced from the MCQ split of the BLEnD dataset\footnote{\url{https://huggingface.co/datasets/nayeon212/BLEnD}, `multiple-choice-questions' split.} \citep{NEURIPS2024_8eb88844}. Our current curation focuses on English to enable controlled comparisons across modalities and leverage state-of-the-art generation models.

\subsection{Dataset Extension Pipeline}
\label{sec:extension_pipeline}
The pipeline proceeds through the following stages:

\textbf{Step 1: MCQ Answer Consolidation.} To associate each base SAQ template with its verified answer set across cultures, we processed the BLEnD MCQ dataset ($\sim$306k instances). For each MCQ instance, we extracted the correct answer text and its corresponding region (`\textit{country}' field). This answer text was subsequently added to the known answers set for the specific question template `\textit{ID}' and region, effectively consolidating all unique correct answers from the MCQ data for each base template across the 16 regions. Invalid answers (e.g., ``\textit{idk}'' or ``\textit{i don't know}'') were excluded.

\textbf{Step 2: Tangibility Filtering.} To ensure that questions are suitable for transformation into a VQA format, we automatically assessed the SAQ templates. Using GPT-4o (\citeyear{HelloGPT4o2024}) (prompt in Figure~\ref{fig:tangibility_prompt}, Appendix~\ref{appendix:prompts}), we classified templates based on whether the question entailed answers that are \textbf{concrete, visually representable entities}. This filtering step selected 313\footnote{Initially, 320 templates were selected. However, during MCQ generation in Step 6, 7 templates lacked sufficient semantically distinct options to form valid 4-option MCQs, resulting in 313 usable templates.} tangible question templates for further processing. For instance, in the example illustrated in Figure \ref{fig:overview}, the answer ``\textit{Rice Pudding}'' is a visually representable entity.

\textbf{Step 3: Question Rephrasing \& Placeholder Generation.} To create the textual condition for testing robustness to linguistic variation, we inverted the standard query format (Region $\rightarrow$ Entity) to (Entity $\rightarrow$ Region). For each of the 313 tangible question templates, GPT-4o was prompted (Prompt in Figure~\ref{fig:rephrasing_prompt}, Appendix~\ref{appendix:prompts}) to generate a canonical rephrased question template. Concurrently, the model generated a generic \textbf{image placeholder} (e.g., `\textit{this food}') designed to replace the entity name in the VQA format, thereby compelling models to rely primarily on the visual modality for that task.

\textbf{Step 4: Image Generation} To enable multimodal evaluations, we generated 4,916 culturally-contextualised images, one for each unique answer-region pair from the 313 tangible templates. We used Gemini 2.5 Flash Image~\citep{fortinIntroducingGemini252025} for image generation,  with prompts conditioned on the original question, specific answer, and region (prompt in Figure~\ref{fig:image_gen_prompt}, Appendix~\ref{appendix:prompts}). To validate the use of synthetic images, we conducted a comparative study across three conditions (Old Synthetic, New Synthetic, and Human-Curated). As detailed in Appendix D.3, model performance on Gemini 2.5 Flash images showed a negligible difference (-1.7\%) from human-curated real-world images, justifying their use as a scalable, high-fidelity proxy.


\textbf{Step 5: Human Validation.} A multi-stage validation process ensured the quality of all generated assets. First, three human annotators assessed the quality and semantic fidelity of 313 rephrased question templates, leading to the manual correction of 39 SAQ templates flagged by majority vote (see Appendix~\ref{appendix:annotation_details} for validation results, and Table~\ref{tab:appendix_rephrase_examples} for examples of such corrections). Second, to validate the quality of the new Gemini 2.5 Flash image set, we designed a rigorous, sampled quality assurance protocol. A stratified random sample of 500 images ($\sim$10\%) is evaluated by three annotators based on conceptual plausibility and recognisability (with instructions to use search engines to verify unfamiliar cultural concepts). This practical approach provides a statistical measure of dataset quality while remaining scalable. Full guidelines and validation details are provided in Appendix \ref{appendix:annotation_details}.


\textbf{Step 6: Parallel MCQ Generation.} To create the final dataset, we generated three parallel MCQ sets for each core fact (validated tangible template + answer/region pair + corresponding image). For each fact, we generated up to 5 unique MCQ instances by sampling semantically distinct distractors (answers from different regions for the same template,  with a simple substring check used to filter overly similar options). Uniqueness was enforced by tracking the set of answer options (1 correct, 3 distractors) for each fact and discarding duplicate sets of options. This yielded three parallel formats that tested the same knowledge point:

\textbf{Original MCQ $O$:} A text-only format querying cultural knowledge in the standard Region $\rightarrow$ Entity form.

\textbf{Rephrased MCQ $R$:} A linguistically inverted Entity $\rightarrow$ Region format testing sensitivity to phrasing variation.

\textbf{VQA-Style MCQ $V$:} A visual-grounded format pairing images with rephrased questions (Image + Placeholder $\rightarrow$ Region).

Examples of these parallel MCQ formats are provided in Appendix~\ref{appendix:mcq_examples} (Table~\ref{tab:parallel_mcq_examples_transposed}). To ensure the validity of the benchmark, a post-hoc uniqueness verification was performed to filter out any parallel set of MCQ instances where a distractor region could also be a valid answer for the queried entity in the $R$ and $V$ formats. This structured generation ensures diverse and non-repetitive test cases for robust cross-modal evaluation.



\subsection{Resulting Dataset and Statistics}
\label{sec:dataset_stats}

\begin{table}[t!]
\small
\centering
\renewcommand{\arraystretch}{0.8}
\setlength{\tabcolsep}{4pt} 
\begin{tabular}{@{} l S[table-format=4.0] S[table-format=3.2, table-space-text-post={\,\%}] @{}} 
\toprule
\textbf{Category} & {\textbf{MCQ Count}} & {\textbf{Percentage}} \\ 
\midrule
\multicolumn{3}{@{}l}{\textit{Breakdown by Topic}} \\
\cmidrule(r){1-1} 
Education                     & 1765 &  8.10\,\% \\
Family                        & 2312 & 10.61\,\% \\
Food                          & 6681 & 30.67\,\% \\
Holidays/Celebration/Leisure & 4294 & 19.71\,\% \\
Sport                         & 4650 & 21.35\,\% \\
Work life                     & 2080 &  9.55\,\% \\
\midrule
\multicolumn{3}{@{}l}{\textit{Breakdown by Country/Region}} \\
\cmidrule(r){1-1} 
Algeria (DZ)              & 1174 &  5.39\,\% \\
Assam (AS)                & 1761 &  8.08\,\% \\
Azerbaijan (AZ)           & 1180 &  5.42\,\% \\
China (CN)                & 1497 &  6.87\,\% \\
Ethiopia (ET)             & 1450 &  6.66\,\% \\
Greece (GR)               & 1449 &  6.65\,\% \\
Indonesia (ID)            & 1451 &  6.66\,\% \\
Iran (IR)                 & 1331 &  6.11\,\% \\
Mexico (MX)               & 1522 &  6.99\,\% \\
North Korea (KP)          & 1287 &  5.91\,\% \\
Northern Nigeria (NG)\tablefootnote{The lower instance count for Northern Nigeria is inherited from the upstream BLEnD data, which had higher rates of annotator 'I don't know' responses and refusals for this region.}           &  998 &  4.58\,\% \\
South Korea (KR)          & 1532 &  7.03\,\% \\
Spain (ES)                & 1398 &  6.42\,\% \\
UK (GB)                   & 1260 &  5.78\,\% \\  
US                        & 1296 &  5.95\,\% \\
West Java (JB)            & 1196 &  5.49\,\% \\
\midrule
\textbf{Total Instances} & \textbf{21,782} & \textbf{100.00\,\%} \\ 
\bottomrule
\end{tabular}
\caption{\textbf{\texttt{BLEnD-Vis}}: MCQ Instance Count and Percentage Breakdown.}
\label{tab:mcq_combined_percent_breakdown}
\end{table}

The final \textbf{\texttt{BLEnD-Vis}} benchmark comprises 313 tangible question templates drawn from BLEnD, each paired with a rephrased version and a corresponding image placeholder to enable controlled evaluation across three modalities. In total, we generated 4,916 culturally grounded images, each corresponding to a unique answer-region pair. The question templates and culturally grounded images were used to construct 21,782 MCQ instances for each of the three parallel formats; each with the same topic and region distribution: \textit{Original}, \textit{Rephrased}, and \textit{VQA-style}. 


The three parallel formats of MCQs enable direct comparison of model performance on the same underlying cultural knowledge presented through different textual phrasings and modalities. For evaluations of cross-modal knowledge via unimodal training, the dataset is further split into training and test sets based on question templates to prevent data leakage (details in Appendix~\ref{appendix:split_details}, Table~\ref{tab:appendix_split_stats}).


%% file: sections/4_results.tex
\section{Results \& Analysis}
\label{sec:4:results}

\begin{table*}[!ht]
\footnotesize 
\centering
\renewcommand{\arraystretch}{0.8}
\setlength{\tabcolsep}{3pt} 
\begin{tabular}{@{} l S[table-format=2.2] S[table-format=2.2] S[table-format=2.2] S[table-format=2.2] S[table-format=2.2] @{}}
\toprule
\textbf{Model} & {\textbf{Original MCQ}} & {\textbf{Rephrased MCQ}} & {\textbf{VQA MCQ}} & {\textbf{R-V Agree (\%)}} & {\textbf{R-V Correct (\%)}} \\ 
\midrule
GPT-4o (\citeyear{HelloGPT4o2024})                      & \textbf{69.56} & \textbf{63.36} & \textbf{92.01} & \underline{66.29} & \textbf{60.83} \\
Qwen2.5-VL-32B (\citeyear{baiQwen25VLTechnicalReport2025})       & \underline{61.90}    & \underline{57.32}    & \underline{86.03}       & 63.83          & \underline{53.59} \\
Kimi-VL-2.8B (\citeyear{teamKimiVLTechnicalReport2025})           & 57.13          & 52.22          & 83.21          & 61.59          & 48.51 \\
Llama-3.2-Vision-11B (\citeyear{Llama32Revolutionizing2024})     & 58.45          & 54.57          & 81.24          & 60.21          & 48.01 \\
Qwen2.5-VL-7B (\citeyear{baiQwen25VLTechnicalReport2025})        & 58.05 & 49.06 & 84.91 & 58.19 & 46.08 \\
Molmo-7B-D (\citeyear{deitkeMolmoPixMoOpen2024})                & 53.99          & 50.57          & 72.39          & 62.82          & 42.89 \\
InternVL3-8B (\citeyear{zhuInternVL3ExploringAdvanced2025})      & 57.18          & 54.68          & 64.07          & 65.79          & 42.27 \\
LLaVA-1.6-13B (\citeyear{leeLLaVANeXTImprovedReasoning2024})     & 44.89          & 50.85          & 67.65          & 63.56          & 41.03 \\
LLaVA-1.6-7B (\citeyear{leeLLaVANeXTImprovedReasoning2024})      & 41.40          & 46.05          & 63.22          & 65.53 & 37.40 \\
PaliGemma2-10B (\citeyear{steinerPaliGemma2Family2024})          & 54.26          & 52.64          & 54.35          & \textbf{67.37} & 37.18 \\
DeepSeek-VL2-small-2.8B (\citeyear{wuDeepSeekVL2MixtureofexpertsVisionlanguage2024}) & 50.76 & 49.43 & 45.95 & 54.40 & 24.89 \\
NVILA-2B (\citeyear{liuNVILAEfficientFrontier2025}) & 40.10          & 43.58          & 42.78          & 60.77          & 23.57 \\
\midrule
\textbf{Mean (Overall)}                                      & \textbf{53.97} & \textbf{52.03} & \textbf{69.82} & \textbf{62.53} & \textbf{42.19} \\
\bottomrule
\end{tabular}
\caption{Zero-Shot Accuracies (\%) of VLMs on \textbf{\texttt{BLEnD-Vis}} (Full Dataset). `R-V Agree \%' indicates the percentage of instances where the model's prediction for the \textbf{Rephrased} MCQ matched its prediction for the \textbf{VQA} MCQ. `R-V Correct \%' indicates the percentage of instances where the model answered \textit{both} the \textbf{Rephrased} and \textbf{VQA} MCQs correctly. Models ordered by `R-V Correct \%'. Best \textbf{bolded}, second best \underline{underlined}.}
\label{tab:main_vlm_results} 
\end{table*}

We evaluated 12 VLMs on the \textbf{\texttt{BLEnD-Vis}} benchmark to assess their robustness in representing everyday cultural knowledge. All evaluations were conducted in a zero-shot setting using the full dataset of 21,782 MCQ instances across three aligned formats. Models were prompted using a standardised evaluation template (Appendix~\ref{appendix:prompts}, Figure~\ref{fig:evaluation_prompt_template}). Table~\ref{tab:main_vlm_results} reports each model's accuracy on the three individual formats and includes two cross-modal consistency metrics. In particular, we sort models by their performance on the `R-V Correct \%' metric, which captures the percentage of cultural facts where the model correctly answered both the rephrased MCQ and the corresponding VQA-style MCQ, highlighting its ability to generalise consistently across modalities.


\subsection{Overall Model Performance}
\label{sec:overall_model_perf}

Table~\ref{tab:main_vlm_results} presents model performance across the three MCQ formats, revealing several notable trends in cultural knowledge robustness and multimodal reasoning.

\textbf{Model size does not consistently predict performance.} While performance generally scales within a model family (e.g., Qwen2.5-VL-32B outperforms the 7B variant on key consistency metrics like `R-V Correct \%`), performance across different model families does not strictly correlate with parameter count. For instance, smaller models like Kimi-VL-2.8B (48.51\% `R-V Correct \%`) and Llama-3.2-Vision-11B (48.01\%) outperform the larger LLaVA-1.6-13B (41.03\%). This suggests that factors beyond sheer parameter scale, such as the diversity of pre-training data, architectural choices for multimodal integration, and specific fine-tuning strategies, play a significant role in encoding robust cultural knowledge.

\textbf{Rephrasing questions slightly reduces performance.}  
Across models, average accuracy declines from 53.97\% on the Original MCQ format to 52.03\% on the Rephrased MCQ format.  This reduced performance may stem from the prevalence of the (Region~$\rightarrow$~Entity) format prevalent across many cultural benchmarks \citep{NEURIPS2024_8eb88844, chiuCulturalBenchRobustDiverse2025, romeroCVQACulturallydiverseMultilingual2024}. This implies that standard benchmark formats may overestimate a model's capability of cultural understanding, as the Entity~$\rightarrow$~Region format may disrupt these learned patterns.


\textbf{Visual input provides important cultural cues.}
Models perform better on the VQA format (69.82\%)  than on both the Rephrased text-only format (52.03\%) and the Original text-only format (53.97\%). For instance,  Kimi-VL-2.8B improves from 52.22\% (Rephrased) to 83.21\%(VQA), and Qwen2.5-VL-7B from 49.06\% to 84.91\%. This is likely because images can convey additional culture-specific cues for cultural-knowledge retrieval while textual prompts might be informationally sparse. However, this implies that evaluating cultural capabilities through VQA alone can be misleading, as high VQA scores may not reflect true multimodal reasoning and may mask underlying weaknesses under unimodal text-only settings.


\textbf{Cross-modal consistency remains a challenge.}  
Despite improved accuracy in the VQA format, models struggle to produce consistent, correct answers across modalities. The average `R-V Agree \%', the proportion of matched predictions across the Rephrased and VQA formats, is moderately high at 62.53\%, but the stricter `R-V Correct \%', accuracy on both formats simultaneously, drops to 42.19\%. Even the top-performing model (GPT-4o) achieves only 60.83\% on `R-V Correct \%'. This gap highlights that many correct answers in one modality are not replicated in the other, reflecting cross-modal fragility. This motivates future work on modal consistency, highlighting the need for benchmarks, evaluations, and training processes that optimise for robust cross-modal grounding over unimodal accuracy.


\subsection{Performance Variation by Region}
\label{sec:region_perf}
\begin{figure*}[htbp]
    \centering
    \includegraphics[width=\linewidth]{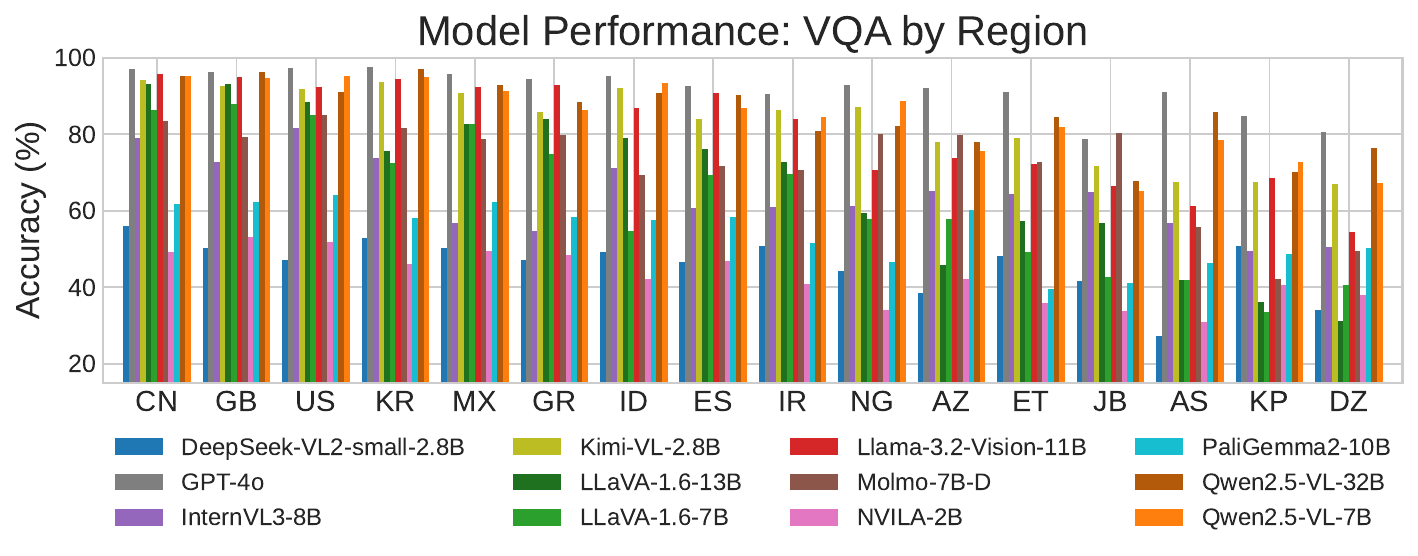}
    \caption{Accuracies (\%) of each evaluated VLM for the VQA-Style MCQ format in \textbf{\texttt{BLEnD-Vis}} (Full Dataset) across 16 different cultural regions (see Appendix~\ref{appendix:region_codes}, Table~\ref{tab:appendix_region_code_map} for region code definitions), highlighting regional variations in model performance. (Original and Rephrased text-only formats are in Appendix~\ref{appendix:region_model_text_performance}.)}
    \label{fig:vqa_region_model_performance}
\end{figure*}

Table~\ref{tab:region_performance_breakdown_main} reports mean accuracy across all models for each cultural region, aggregated over the three MCQ formats. Figure~\ref{fig:vqa_region_model_performance} visualises VQA-style performance per region and model, highlighting distinct regional strengths and weaknesses. Additional breakdowns for the Original and Rephrased text-only formats are included in Appendix~\ref{appendix:region_model_text_performance}.

\begin{table}[!ht] 
\footnotesize
\centering
\renewcommand{\arraystretch}{0.8}
\setlength{\tabcolsep}{2pt}
\begin{tabular}{@{} l S[table-format=2.2] S[table-format=2.2] S[table-format=2.2] S[table-format=2.2] @{}}
\toprule
\textbf{Region} & {\textbf{Original}} & {\textbf{Rephrased}} & {\textbf{VQA}} & {\textbf{Mean}} \\
\midrule
US                    & 64.92 & \textbf{77.42} & 80.87 & \textbf{74.40} \\
UK (GB)               & \textbf{65.97} & 69.95 & 81.15 & 72.36 \\
China (CN)            & 61.87 & 67.76 & \textbf{82.20} & 70.61 \\
South Korea (KR)      & 58.56 & 66.99 & 78.18 & 67.91 \\
Mexico (MX)           & 59.12 & 55.00 & 77.11 & 63.74 \\
Spain (ES)            & 58.17 & 55.20 & 72.85 & 62.07 \\
Indonesia (ID)        & 52.84 & 59.01 & 73.48 & 61.78 \\
Greece (GR)           & 54.39 & 48.42 & 74.57 & 59.13 \\
Iran (IR)             & 52.13 & 46.27 & 70.27 & 56.22 \\
Azerbaijan (AZ)       & 54.13 & 43.47 & 65.56 & 54.39 \\
Northern Nigeria (NG) & 44.41 & 48.33 & 67.10 & 53.28 \\
West Java (JB)        & 44.16 & 44.29 & 59.27 & 49.24 \\
Ethiopia (ET)         & 43.79 & 37.45 & 64.67 & 48.64 \\
Assam (AS)            & 46.63 & 37.97 & 57.07 & 47.22 \\
Algeria (DZ)          & 50.74 & 36.75 & 53.34 & 46.94 \\
North Korea (KP)      & 49.46 & 35.48 & 55.46 & 46.80 \\
\midrule
\textbf{Mean (Regions)} & \textbf{53.83} & \textbf{51.86} & \textbf{69.57} & \textbf{58.42} \\
\bottomrule
\end{tabular}
\caption{Model Accuracies (\%) by Country/Region on \textbf{\texttt{BLEnD-Vis}} (Full Dataset), ordered by mean task performance.}
\label{tab:region_performance_breakdown_main}
\end{table}

\textbf{Model performance varies significantly by region, often reflecting resource disparities.}  
Regions with greater representation in publicly available training data tend to yield higher model performance. For example, the US, UK, and China achieve average accuracies of \textbf{74.40\%}, \textbf{72.36\%}, and \textbf{70.61\%}, respectively. In contrast, less digitally represented regions such as North Korea, Algeria, and Assam score markedly lower, with mean accuracies of \textbf{46.80\%}, \textbf{46.94\%}, and \textbf{47.22\%}. This trend aligns with prior findings from BLEnD~\citep{NEURIPS2024_8eb88844} and underscores persistent gaps in cultural representation within pretraining corpora~\citep{taoCulturalBiasCultural2024}.

\textbf{The (Entity $\rightarrow$ Region) query format exacerbates performance gaps between high and low-resource regions.}
For the \textbf{Original MCQ}, the performance range between the top-performing region (UK: \textbf{65.97\%}) and a bottom-performing region (e.g., Ethiopia: \textbf{43.79\%}) is approximately 22\%. This gap significantly widens for the \textbf{Rephrased MCQ} (US: \textbf{77.42\%} vs. North Korea: \textbf{35.48\%}) and the VQA format (China: \textbf{82.20\%} vs. Algeria: \textbf{53.34\%}). This suggests that when cued with an entity (textually or visually) and asked for its associated region, models may default to high-resource regions if their knowledge of the entity's specific origin in a low-resource region is less robust. This suggests that the (Entity $\rightarrow$ Region) format, particularly in VQA, is a more discriminative test of potential regional biases and deep cultural grounding. For instance, as illustrated in Figure~\ref{fig:vqa_region_model_performance}, models like Qwen2.5-VL-7B exhibit strong VQA performance for China (95.26\%), while their scores for a region like Algeria are much lower (67.38\% ). Conversely, models like PaliGemma2-10B and Molmo-7B-D, while generally strong for US VQA (64.04\% and 85.03\% respectively), show comparatively lower performance for China (61.72\% and 83.58\% respectively).

\subsection{Cross-Modal Knowledge Transfer}
\label{sec:cross_modal_transfer}

To explore whether cultural knowledge learned in one modality can transfer to another, we conducted preliminary cross-modal fine-tuning experiments on a subset of models. In the first setting (Text-trained~$\rightarrow$~VQA-test), models were fine-tuned on the training split of the Rephrased Text-Only MCQs and evaluated on the test split of the VQA-Style MCQs. In the reverse setting (VQA-trained~$\rightarrow$~Text-test), models were fine-tuned on VQA-Style MCQs and evaluated on Rephrased Text-Only MCQs. In both cases, we used the same 80/20 template-based train/test split to prevent leakage of cultural facts between training and evaluation (Appendix~\ref{appendix:split_details}). Full fine-tuning hyperparameters are provided in Appendix~\ref{appendix:finetuning_details}.

\begin{table}[!ht]
\small
\centering
\renewcommand{\arraystretch}{0.4}
\setlength{\tabcolsep}{5pt}
\begin{tabular}{@{} l S[table-format=2.2] l @{}}
\toprule
\multicolumn{3}{@{}l}{\textbf{Target Task: Rephrased Text (Test Set)}} \\ 
\cmidrule(r){1-3}
\multirow{2}{*}{\textbf{Model}} & \multicolumn{2}{c}{\textbf{Performance (\%)}} \\ 
\cmidrule(lr){2-3}
& {\textbf{Baseline}} & {\textbf{VQA-Trained}} \\ 
\midrule
LLaVA-1.6-7B         & 44.22 & 52.82 (+8.60\,\%) \\
Qwen2.5-VL-7B        & 47.83 & 51.66 (+3.83\,\%) \\
PaliGemma2-10B       & 51.39 & 53.72 (+2.33\,\%) \\
Llama-3.2-Vision-11B & 51.73 & 52.91 (+1.18\,\%) \\
\midrule
\textbf{Mean (Models)} & \textbf{48.79} & \textbf{52.78 (+3.98\,\%)} \\ 
\bottomrule 
\\
\toprule 
\multicolumn{3}{@{}l}{\textbf{Target Task: VQA (Test Set)}} \\ 
\cmidrule(r){1-3}
\multirow{2}{*}{\textbf{Model}} & \multicolumn{2}{c}{\textbf{Performance (\%)}} \\ 
\cmidrule(lr){2-3}
& {\textbf{Baseline}} & {\textbf{Text-Trained}} \\ 
\midrule
LLaVA-1.6-7B         & 63.63 & 78.35 (+14.72\,\%) \\
Qwen2.5-VL-7B        & 85.25 & 91.06 (+5.81\,\%) \\
PaliGemma2-10B       & 52.67 & 92.60 (+39.93\,\%) \\
Llama-3.2-Vision-11B & 80.64 & 84.78 (+4.14\,\%) \\
\midrule
\textbf{Mean (Models)} & \textbf{70.55} & \textbf{86.70 (+16.15\,\%)} \\ 
\bottomrule
\end{tabular}
\caption{Cross-Modal Transfer Performance (\% Accuracy on Test Set). `Baseline' refers to zero-shot performance on the target task. `VQA-Trained' and `Text-Trained' refer to performance on the target task after fine-tuning on VQA or Rephrased Text training data, respectively. Percentage improvement over baseline shown in parentheses.}
\label{tab:cross_modal_transfer_results} 
\end{table}

Table~\ref{tab:cross_modal_transfer_results} summarises these transfer learning results, detailing performance on the target task after fine-tuning on the source task, compared to their zero-shot baselines. The results indicate varying degrees of knowledge transfer across modalities.

Notably, fine-tuning on the Rephrased Text-Only MCQs consistently leads to improved performance on the VQA-Style MCQs (Text-trained~$\rightarrow$~VQA-test). All four evaluated models show gains in this direction: LLaVA-1.6-7B (+14.72\%), Qwen2.5-VL-7B (+5.81\%), PaliGemma2-10B (+39.93\%), and Llama-3.2-Vision-11B (+4.14\%). On average, text-based training boosted VQA performance by \textbf{+16.15\%}. This suggests that strengthening the model's textual understanding of the (Entity $\rightarrow$ Region) relationship and associated cultural facts can positively transfer to its ability to perform the same task when presented with visual cues, potentially by solidifying semantic representations that the visual modality can then leverage more effectively.

Conversely, the transfer from VQA training to Rephrased Text-Only performance (VQA-trained~$\rightarrow$~Text-test) is more modest: with LLaVA-1.6-7B (+8.60\%), Qwen2.5-VL-7B (+3.83\%), PaliGemma2-10B (+2.33\%), and Llama-3.2-Vision-11B (+1.18\%) all showing lower improvements.  The average improvement across these models is modest at \textbf{+3.98\%}. This might suggest that while VQA training exposes models to visual-textual pairings of cultural concepts, it may not consistently enhance (and could potentially interfere with) purely textual reasoning pathways. It is plausible that VQA training could lead to an over-reliance on visual features or that the VQA task structure does not reinforce nuanced textual understanding as effectively as direct textual training.

These findings highlight the interplay between modalities in representing and reasoning about cultural knowledge. The positive transfer from text to VQA is promising, suggesting robust textual understanding as foundational. However, the less consistent transfer from VQA to text warrants further investigation into how multimodal training influences distinct reasoning pathways.

\begin{table}[t!]
\small \centering
\renewcommand{\arraystretch}{0.4} \setlength{\tabcolsep}{4pt}
\begin{tabular}{@{} l S[table-format=2.2] S[table-format=2.2] S[table-format=2.2] S[table-format=2.2] @{}}
\toprule
\textbf{Topic} & {\textbf{Original}} & {\textbf{Rephrased}} & {\textbf{VQA}} & {\textbf{Mean}} \\ 
\midrule
Work life          & \textbf{67.34} & \textbf{66.04} & 71.15 & \textbf{68.18} \\
Holidays/Celeb.    & 56.15 & 55.18 & \textbf{71.38} & 60.90 \\
Sport              & 55.32 & 52.04 & 71.28 & 59.55 \\
Food               & 52.90 & 50.27 & 67.14 & 56.77 \\
Education          & 45.90 & 44.57 & 70.67 & 53.71 \\
Family             & 44.44 & 44.31 & 69.88 & 52.88 \\
\midrule
\textbf{Mean} & \textbf{53.68} & \textbf{52.07} & \textbf{70.25} & \textbf{58.66} \\
\bottomrule
\end{tabular}
\caption{Mean Model Performance (\%) by Topic on \textbf{\texttt{BLEnD-Vis}} (Full Dataset).}
\label{tab:appendix_topic_performance_table}
\end{table}

\subsection{Performance Variation by Topic}
The distribution of mean model performance by topic (Table~\ref{tab:appendix_topic_performance_table}) reveals substantial variation in task difficulty across cultural domains. Models performed best on `\textit{Work life}` (mean accuracy: 68.18\%) and `\textit{Holidays/Celebration/Leisure}` (60.90\%), while topics such as `\textit{Family}` (52.88\%) and `\textit{Education}` (53.71\%) posed greater challenges. These differences may reflect varying levels of specificity, visual distinctiveness, or semantic ambiguity associated with the entities in each topic.

Across all categories, performance was consistently higher on the VQA format compared to the Rephrased text-only format. For example, accuracy on `Education` increased from \textbf{44.57\%} (Rephrased) to \textbf{70.67\%} (VQA), and on `\textit{Family}` from \textbf{44.31\%} to \textbf{69.88\%}. These results suggest that visual input provides a valuable disambiguating signal, particularly for topics where textual rephrasings may be less canonical or culturally entangled. The consistent VQA boost underscores the utility of grounded visual context in supporting flexible retrieval of everyday cultural knowledge.


%% file: sections/5_discussion.tex
\section{Discussion}
\label{sec:5:discussion}
Our findings highlight several challenges in current VLMs' representation of everyday cultural knowledge. First, the consistent drop in accuracy under linguistic rephrasing suggests that many models rely on superficial pattern matching rather than robust conceptual understanding. While visual input in the VQA format generally improves performance, low cross-modal consistency (particularly in joint correctness) reveals a lack of integration between textual and visual representations.

Regional disparities further underscore equity concerns: models perform significantly worse on queries involving lower-resource regions, a gap potentially exacerbated by limitations in VLM training data and the cultural fidelity of generated images. This motivates greater inclusivity in pretraining data and culturally-aware generative tools.

Notably, model scale does not reliably predict performance. Instead, our results suggest that the diversity and specificity of pretraining data, along with architectural design, may be more critical to cultural robustness. The observed asymmetry in cross-modal transfer, where text-based fine-tuning enhances VQA performance but not vice versa, reinforces the foundational role of linguistic grounding in multimodal understanding.

Together, these insights call for a shift in VLM development: beyond factual recall and scale, toward deeper, transferable, and culturally representative knowledge across modalities. Future training strategies could explicitly up-weight underrepresented cultural samples to counteract high-resource bias, and include structurally diverse templates (such as inverted Entity $\rightarrow$ Region formats) to improve linguistic robustness.

%% file: sections/6_conclusion.tex
\section{Conclusion \& Future Work}
\label{sec:6:conclusion}

We introduced \textbf{\texttt{BLEnD-Vis}}, a multimodal benchmark for evaluating the robustness and visual grounding of everyday cultural knowledge in vision-language models. Covering 16 culturally diverse regions, the benchmark includes 313 question templates, 4,916 generated images, and over 21,000 MCQ instances across three formats. Evaluations of 12 VLMs reveals key limitations: (i) performance degrades under linguistic rephrasing, cross-modal consistency remains low, and regional disparities persist for underrepresented cultures. (ii) Model scale does not reliably predict success, while fine-tuning results suggest that strong textual grounding supports more effective visual transfer. Future work includes expanding to multilingual settings, analysing failure patterns, improving culturally-aware training and generation methods, and extending evaluations to open-ended tasks.



%% file: sections/7_limitations.tex
\section*{Limitations}
\label{sec:7:limitations}

While \textbf{\texttt{BLEnD-Vis}} provides a novel framework for evaluating cultural robustness in VLMs, several limitations should be acknowledged. First, the benchmark is constructed entirely in English, limiting its applicability to multilingual and cross-lingual settings and inherently framing cultural concepts through an Anglophone lens. Future extensions should explore culturally grounded evaluation in other languages and code-mixed contexts. Second, the VQA component relies on synthetically generated images. Although human validation was performed, these images may still exhibit subtle inaccuracies or stereotypical cues, influenced by the biases of the image generation models. This could affect the fairness and fidelity of the VQA evaluation, particularly for underrepresented cultures. Furthermore, while our validation confirms synthetic images are a valid performance proxy, they may "flatten" cultural nuances compared to real artifacts (e.g., depicting generic rather than region-specific variants). Third, \textbf{\texttt{BLEnD-Vis}} focuses on tangible, everyday cultural knowledge. More abstract cultural dimensions, such as values, norms, or social rituals—are not represented, leaving a gap in evaluating deeper forms of cultural competence. Fourth, the use of MCQs limits evaluation to discriminative reasoning. While suitable for controlled comparisons, MCQs do not capture generative abilities such as explaining cultural facts, expressing empathy, or engaging in culturally appropriate open-ended dialogue. As such, the benchmark does not reflect models' full potential in real-world, interactive scenarios. Fifth, the current VQA format (Image + Placeholder $\rightarrow$ Region) tests only one type of visual grounding. Alternative visual query structures could expose different strengths or weaknesses in multimodal reasoning. Lastly, while human validation was conducted, annotator familiarity may have varied across the 16 regions. For less globally prominent cultures, subtle inaccuracies or overlooked errors may persist. These limitations point to important directions for future work, including multilingual expansion, generative evaluation, culturally adaptive image synthesis, and a broader model evaluation landscape.

%% file: sections/8_ethical_statement.tex
\section*{Ethics Statement}
\label{sec:8:ethical_statement}

The development and deployment of \textbf{\texttt{BLEnD-Vis}} were guided by a commitment to responsible research practices. We acknowledge several ethical considerations inherent in evaluating cultural knowledge in AI models.

\textbf{Bias and Representation:} The benchmark aims to \textit{reveal} potential biases in VLMs concerning cultural knowledge, particularly disparities between high-resource and lower-resource regions, as evidenced in our results. However, the benchmark itself could inadvertently perpetuate biases present in the original BLEnD data or introduced during image generation (e.g., stereotypical depictions), despite human validation efforts. By making the benchmark public, we intend to facilitate research into identifying and mitigating such biases, promoting more equitable cultural representation in future models. We focused on everyday cultural knowledge to minimise the risk of evaluating sensitive or sacred topics inappropriately.

\textbf{Data Provenance and Annotation:} The textual data originates from the BLEnD dataset \citep{NEURIPS2024_8eb88844}, which involved human participants providing cultural information. Consistent with the original BLEnD dataset's licensing, we release our dataset under a CC-BY-4.0 license. Images were generated using publicly available models and subsequently validated by human annotators following defined guidelines (Appendix \ref{appendix:annotation_details}) to ensure relevance and appropriateness, filtering out problematic content flagged by a majority vote. Annotators are focused on validation rather than subjective judgment of cultural value. Furthermore, AI assistants provided support for coding tasks and enhancing the clarity of manuscript drafts; all such contributions were meticulously reviewed and edited by the authors to ensure the final work's accuracy and adherence to academic standards.

\textbf{Intended Benefit:} Our primary goal is to contribute positively to the AI community by providing a tool that encourages the development of VLMs with a more nuanced, robust, and equitable understanding of diverse global cultures. We believe that improving the cultural competence of AI systems is crucial for fostering user trust, reducing harm caused by cultural insensitivity, and promoting fairness in global AI applications. The dataset and associated code will be released publicly to facilitate transparency and further research in this critical area.

%% file: sections/9_acknowledgement.tex
\section*{Acknowledgements}
\label{sec:9:acknowledgement}
This research project is supported by the National Research Foundation, Singapore, under its National Large Language Models Funding Initiative (AISG Award No: AISG-NMLP-2024-004 and AISG-NMLP-2024-005). Any opinions, findings and conclusions or recommendations expressed in this material are those of the author(s) and do not reflect the views of the National Research Foundation and Ministry of Education, Singapore. We also sincerely thank Yuriel Ryan, Yujia Hu, Dylan Raharja and Zhuoran Wang for their contributions with annotation and evaluation.

%% file: sections/10_appendix.tex
\section{Detailed Performance Breakdowns}
\label{appendix:detailed_performance}

This appendix provides further detailed breakdowns of model performance.

\subsection{Region Code Mapping}
\label{appendix:region_codes}
Table~\ref{tab:appendix_region_code_map} lists the 16 cultural regions included in the \textbf{\texttt{BLEnD-Vis}} benchmark and their corresponding two-letter codes used internally and in some data representations.

\begin{table}[!ht] 
\small
\centering
\renewcommand{\arraystretch}{1.1}
\setlength{\tabcolsep}{6pt}
\begin{tabular}{@{}ll@{}} 
\toprule
\textbf{Code} & \textbf{Country / Region Name} \\
\midrule
AS & Assam \\
AZ & Azerbaijan \\
CN & China \\
DZ & Algeria \\ 
ES & Spain \\
ET & Ethiopia \\
GB & United Kingdom (UK) \\
GR & Greece \\
ID & Indonesia \\
IR & Iran \\
JB & West Java \\
KP & North Korea \\
KR & South Korea \\
MX & Mexico \\
NG & Northern Nigeria \\
US & United States (US) \\
\bottomrule
\end{tabular}
\caption{Mapping of Region Codes to Country/Region Names.}
\label{tab:appendix_region_code_map}
\end{table}

\subsection{Performance by Region and Model (Text-Only Formats)}
\label{appendix:region_model_text_performance}
\begin{figure*}[!ht]
    \centering
    \includegraphics[width=\linewidth]{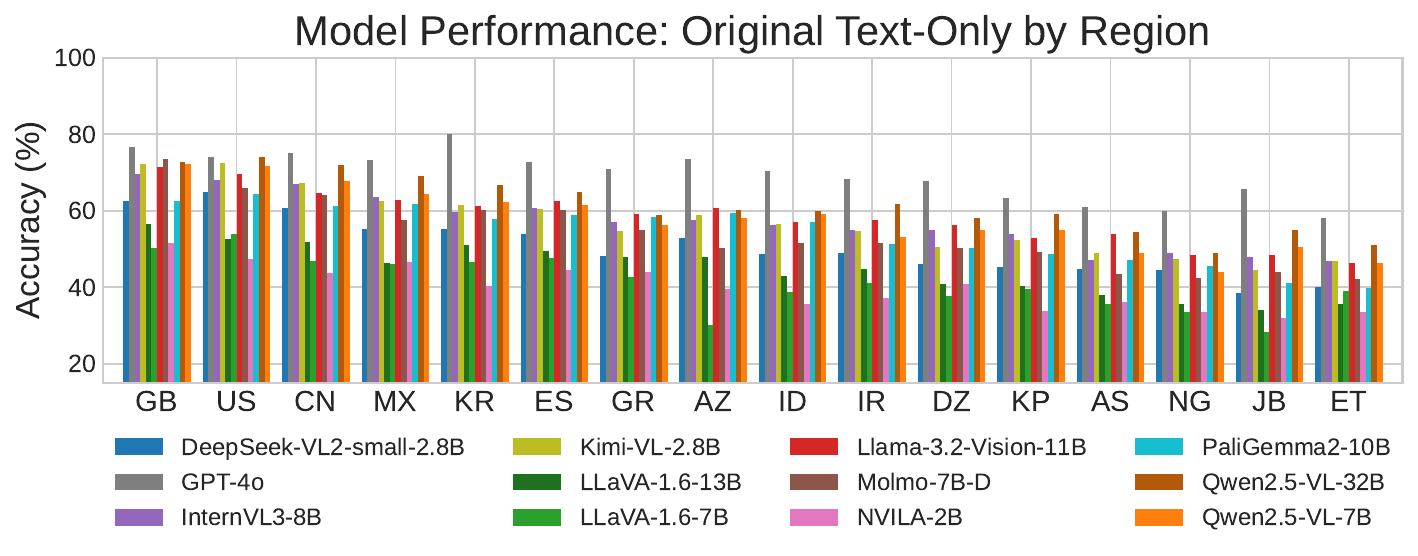}
    \caption{Original Text-Only MCQ Performance (\%) by Region and Model on \textbf{\texttt{BLEnD-Vis}} (Full Dataset).}
    \label{fig:original_region_model_performance}
\end{figure*}

\begin{figure*}[!ht]
    \centering
    \includegraphics[width=\linewidth]{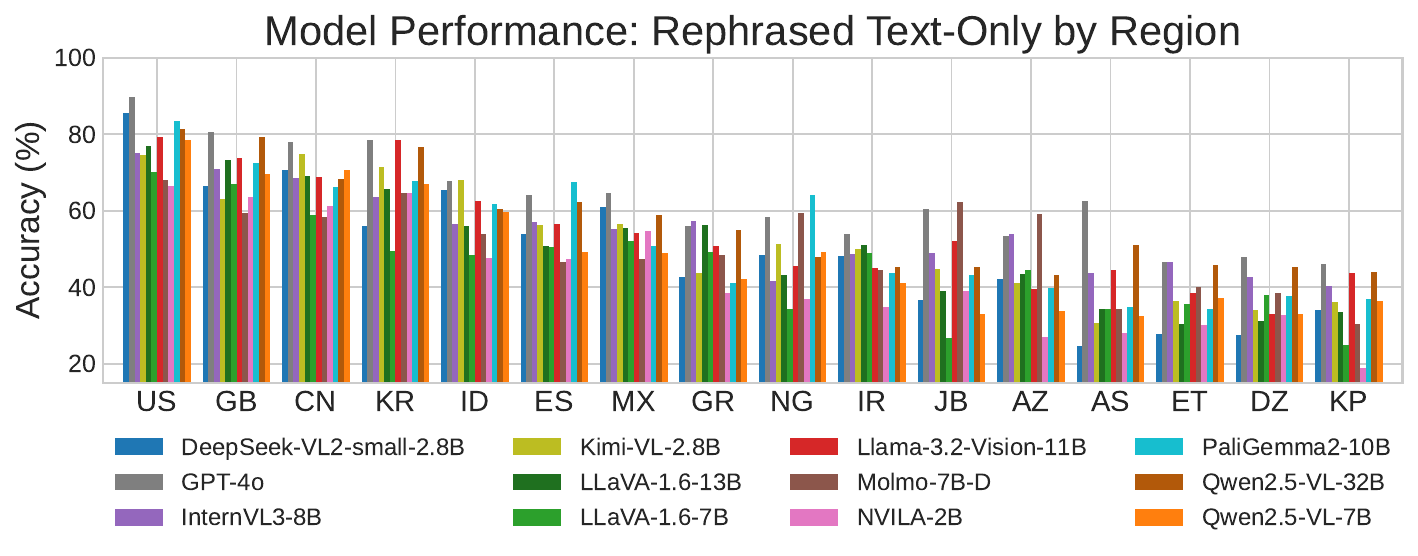}
    \caption{Rephrased Text-Only MCQ Performance (\%) by Region and Model on \textbf{\texttt{BLEnD-Vis}} (Full Dataset).}
    \label{fig:rephrased_region_model_performance}
\end{figure*}

Figures \ref{fig:original_region_model_performance} and \ref{fig:rephrased_region_model_performance} illustrate the performance of each model on the Original Text-Only MCQ and Rephrased Text-Only MCQ formats, respectively, broken down by cultural region.

\section{Dataset Split Details}
\label{appendix:split_details}

The \textbf{\texttt{BLEnD-Vis}} dataset, comprising 21,782 MCQ instances derived from 313 unique question template IDs, was partitioned into training and test sets. The split was performed at the template ID level using an 80\%-20\% ratio (250 IDs for training, 63 IDs for testing) with stratification based on the topic category. This ensures that all MCQ instances originating from the same base template reside in the same split, preventing data leakage. Table~\ref{tab:appendix_split_stats} details the resulting distribution of MCQ instances across topics for the training and test sets.

\begin{table}[htbp]
\small
\centering
\renewcommand{\arraystretch}{0.8}
\setlength{\tabcolsep}{6pt} 

\begin{tabular}{@{} l l S[table-format=5.0] S[table-format=3.1, table-space-text-post={\,\%}] @{}}
\toprule
\textbf{Topic} & \textbf{Split} & {\textbf{Count}} & {\textbf{Percentage}} \\
\midrule
Education & \hspace{1em} Total &  1765 &   8.1\,\% \\ 
          & \hspace{1em} Train &  1366 &   7.9\,\% \\ 
          & \hspace{1em} Test  &   399 &   8.9\,\% \\
\midrule
Family    & \hspace{1em} Total &  2312 &  10.6\,\% \\
          & \hspace{1em} Train &  1823 &  10.5\,\% \\
          & \hspace{1em} Test  &   489 &  11.0\,\% \\
\midrule
Food      & \hspace{1em} Total &  6681 &  30.7\,\% \\
          & \hspace{1em} Train &  5302 &  30.6\,\% \\
          & \hspace{1em} Test  &  1379 &  30.9\,\% \\
\midrule
Holidays/Celeb. & \hspace{1em} Total &  4294 &  19.7\,\% \\ 
                & \hspace{1em} Train &  3512 &  20.3\,\% \\
                & \hspace{1em} Test  &   782 &  17.5\,\% \\
\midrule
Sport     & \hspace{1em} Total &  4650 &  21.3\,\% \\
          & \hspace{1em} Train &  3575 &  20.6\,\% \\
          & \hspace{1em} Test  &  1075 &  24.1\,\% \\
\midrule
Work life & \hspace{1em} Total &  2080 &   9.5\,\% \\
          & \hspace{1em} Train &  1742 &  10.1\,\% \\
          & \hspace{1em} Test  &   338 &   7.6\,\% \\
\midrule \midrule 
\textbf{Overall} & \hspace{1em} Total & \textbf{21782} & \textbf{100.0\,\%} \\
                 & \hspace{1em} Train & \textbf{17320} & \textbf{100.0\,\%} \\
                 & \hspace{1em} Test  &  \textbf{4462} & \textbf{100.0\,\%} \\
\bottomrule
\end{tabular}
\caption{Topic Distribution in Total, Train, and Test Splits of \textbf{\texttt{BLEnD-Vis}} MCQs.}
\label{tab:appendix_split_stats}
\end{table}

\section{Fine-tuning Details for Cross-Modal Transfer Experiments}
\label{appendix:finetuning_details}

This section outlines the experimental setup for the cross-modal transfer learning experiments discussed in Section~\ref{sec:cross_modal_transfer}.

\subsection{Dataset and Splitting}
All fine-tuning and evaluation for the transfer experiments were conducted using the \textbf{\texttt{BLEnD-Vis}} dataset. We utilised the predefined train-test split detailed in Appendix~\ref{appendix:split_details}, which partitions the 313 unique question template IDs into an 80\% training set (250 IDs; 17,320  MCQs) and a 20\% test set (63 IDs; 4,462 MCQs). This split ensures that no underlying cultural facts or question templates seen during training are present in the test set, preventing data leakage.

Two primary training scenarios were investigated:
\begin{enumerate}
    \item \textbf{Text-trained $\rightarrow$ VQA-test:} Models were fine-tuned on the Rephrased Text-Only MCQs from the training split and subsequently evaluated on the VQA-Style MCQs from the test split.
    \item \textbf{VQA-trained $\rightarrow$ Text-test:} Models were fine-tuned on the VQA-Style MCQs from the training split and subsequently evaluated on the Rephrased Text-Only MCQs from the test split.
\end{enumerate}
Zero-shot baseline performance was established by evaluating the pre-trained models directly on the respective test splits without any \textbf{\texttt{BLEnD-Vis}} specific fine-tuning.

\subsection{Fine-tuning Hyperparameters}
The fine-tuning process for both LLaVA-1.6-7B and Qwen2.5-VL-7B utilised a consistent set of key hyperparameters, aiming for a standardised comparison. LoRA \citep{huLoRALowrankAdaptation2021} was employed for efficient fine-tuning, targeting all linear layers. The models were trained for 3 epochs, and the checkpoint corresponding to the best validation loss (or training loss if a separate validation set was not carved out from the training split for hyperparameter tuning) was selected for final evaluation. Table~\ref{tab:finetuning_hyperparams} lists the pertinent hyperparameters.

\subsection{Model Size and Computational Resources}
\label{appendix:computational_resources}
All experiments were conducted utilising NVIDIA H100 GPUs (80GB). For inference, a single GPU with a batch size of 1 was employed. Vision-Question Answering (VQA) tasks required approximately 1.5 to 2.5 hours per model for the full dataset, and around 10 minutes for split dataset evaluations. Text-only task inference was faster, taking 30 to 50 minutes for the full dataset and approximately 10 minutes for split dataset runs per model. Model training was performed on a distributed setup of 8 NVIDIA H100 GPUs, with an average training duration of approximately 1 hour per model configuration.

\begin{table}[!ht]
\small
\centering

\renewcommand{\arraystretch}{1.1}
\begin{tabular}{@{}ll@{}}
\toprule
\textbf{Hyperparameter} & \textbf{Value} \\
\midrule
Fine-tuning Type & LoRA \\
LoRA Rank ($r$) & 8 \\
LoRA Target Modules & All linear layers \\
Learning Rate & 1.0e-4 \\
Number of Train Epochs & 3.0 \\
LR Scheduler Type & Cosine \\
Warmup Ratio & 0.1 \\
Batch Size (per device) & 8 \\
Gradient Accumulation Steps & 8 \\
Mixed Precision & bf16 \\
Optimizer & AdamW \\ 
Weight Decay & 0.01 \\ 
Max Sequence Length & 4000 (cutoff\_len) \\
\bottomrule
\end{tabular}
\caption{Key Fine-tuning Hyperparameters.}
\label{tab:finetuning_hyperparams}
\end{table}

\section{Human Annotation and Image Validation}
\label{appendix:annotation_details} 

To ensure the quality of all generated assets and to validate our use of synthetic images, we conducted a multi-stage human validation process. This involved (1) validating all rephrased question templates, (2) performing sampled quality assurance on the main image dataset, and (3) conducting a direct comparative study of our synthetic images against a human-curated baseline. All annotators are research assistants with at least an undergraduate degree.

\subsection{Rephrased Question Validation}
Three independent annotators evaluated the 320 automatically generated rephrased question templates for semantic fidelity and clarity against the original BLEnD SAQ templates. The goal was to verify if the rephrased question accurately inverted the original query while maintaining clarity.

\textbf{Results:} A majority vote (at least 2 of 3 annotators) flagged 39 templates (12.2\%) as 'Bad', often due to semantic misalignment or grammatical issues. These were manually corrected by the research team. An additional 7 templates were later excluded due to insufficient distractor options, resulting in the 313 validated templates used for the final MCQ dataset.
\subsection{Sampled Quality Assurance for Generated Images}
All 4,916 images in \textbf{\texttt{BLEnD-Vis}} were generated using the Gemini 2.5 Flash model. To quantify the quality of this large dataset, we implemented a sampled quality assurance protocol.

\textbf{Task Setup:} A random, stratified sample of 500 images ($\sim$10\%) was evaluated by three independent annotators. The annotators' goal was to determine if each generated image served as a plausible and recognisable visual representation of its intended cultural concept, based on the entity, region, and a descriptive placeholder.


\textbf{Annotation Guidelines:}
Annotators were provided with a detailed protocol to determine if a generated image serves as a plausible and recognisable visual representation of a specific cultural concept. For each image, annotators were given three key pieces of context: the specific `entity` (e.g., "spicy potatoes"), the `region` (e.g., "Spain"), and the general category or `image\_placeholder` (e.g., "this food").

The core instructions were as follows:

\begin{itemize}
    \item \textbf{Mark as 'G' (Good) if:} The image successfully conveys the intended concept. The guiding principle was whether someone familiar with the region's cultural context would likely understand that the image represents the intended `entity`. Images were marked 'G' even with minor flaws, such as:
        \begin{itemize}
            \item Slightly unusual art styles (e.g., painterly or illustrative), as long as the subject was identifiable.
            \item Minor visual artefacts or imperfections (e.g., odd textures, background glitches).
            \item Representations that might seem generic in isolation but were appropriate within the given cultural context.
        \end{itemize}
    \item \textbf{Mark as 'B' (Bad) if:} The image fails to sufficiently represent or convey the core concept. Annotators were required to provide a brief reason for their judgment. The primary criteria for a 'B' rating were:
        \begin{itemize}
            \item \textit{Clearly Wrong Subject:} The image depicts something fundamentally different from the `entity`.
            \item \textit{Misleading or Ambiguous Representation:} The image is so generic, abstract, or poorly rendered that it fails to represent the `entity`, even with the provided context.
            \item \textit{Unusably Distorted:} The image has such severe visual artefacts that the main subject is unrecognisable or nonsensical.
        \end{itemize}
\end{itemize}
Annotators were also encouraged to use image search engines to verify concepts with which they were unfamiliar, ensuring a high degree of accuracy in their judgments.

\textbf{Results:} Based on a 2/3 majority vote, \textbf{27/500 (5.40\%)} of the sampled images were flagged as 'B' (Bad). This low error rate suggests a strong overall quality and recognisability for the Gemini 2.5 Flash image set, supporting its suitability for our main evaluations.

\subsection{Validation Against Human Curation}
A key concern for benchmarks using synthetic data is whether the data serves as a valid proxy for real-world scenarios. To address this directly, we conducted a controlled experiment comparing VLM performance on our new synthetic images against both an older generation of synthetic images and a newly created, human-curated set of real-world images.

\textbf{Methodology:}
\begin{enumerate}
\item We randomly sampled a set of 100 cultural facts from our benchmark, disjoint from the human quality assurance sample.
\item For these 100 facts, we created three parallel sets of images:
\begin{itemize}
\item \textbf{Synthetic (2.0):} Images generated using an older model (Gemini 2.0 Flash Image).
\item \textbf{Synthetic (2.5):} The final images used in our benchmark, generated by Gemini 2.5 Flash Image.
\item \textbf{Human-Curated:} Real-world images manually sourced by annotators via web search engines to best represent the target concepts. To ensure strict compliance with intellectual property standards, these images were used exclusively for this internal comparative experiment and are not included in the public benchmark release. This allows us to validate performance against real-world data while ensuring the released dataset remains open-source and license-free.
\end{itemize}
\item We ran VQA evaluations for a subset of models across these three parallel image sets.
\end{enumerate}

\textbf{Results and Justification:}
Table~\ref{tab:appendix_image_validation} presents the comparative VQA performance. The results show that \textbf{model performance on the new Gemini 2.5 Flash images is nearly identical to performance on the human-curated images}, with a mean difference of -1.7\%. In contrast, the older synthetic images (Gemini 2.0 Flash) resulted in a significant performance drop of over 21.3\% compared to the human-curated set.
This experiment provides strong evidence that the high-fidelity Gemini 2.5 Flash images serve as a valid proxy for real-world data in our evaluation context, ensuring our results are robust and representative of real-world visual understanding challenges.

\begin{table}[!ht]
\small \centering
\renewcommand{\arraystretch}{1.1} 
\setlength{\tabcolsep}{2pt} 

\newcommand{\stackcell}[1]{\begin{tabular}{@{}c@{}}#1\end{tabular}}

\begin{tabular}{@{} l S[table-format=2.2] c c @{}}
\toprule
\textbf{Model} & {\stackcell{Human \\ Curated}}& {\stackcell{Synth (2.0) \\ (vs. Human)}} & {\stackcell{Synth (2.5) \\ (vs. Human)}} \\
\midrule
Qwen2.5-VL-32B       & 83.00 & 63.00 (-20.0) & 81.00 (-2.0) \\
Llama-3.2-Vision-11B & 78.00 & 55.00 (-23.0) & 76.00 (-2.0) \\
LLaVA-1.6-7B         & 68.00 & 47.00 (-21.0) & 67.00 (-1.0) \\
\midrule
\textbf{Mean (Overall)} & \textbf{76.33} & \textbf{55.00 (-21.3)} & \textbf{74.67 (-1.7)} \\
\bottomrule
\end{tabular}
\caption{VQA performance comparison. Parentheses show the performance change (\% absolute) of synthetic images relative to the human-curated baseline.}
\label{tab:appendix_image_validation}
\end{table}

\section{Annotation Examples: Rephrased Questions and Images}
\label{appendix:annotation_examples}

This section provides illustrative examples of issues identified during the human annotation phase (Task 1: Rephrased Question Validation and Task 2: Image Validation), highlighting common failure modes in automated generation and the importance of the validation step.

\subsection{Examples of Corrected Rephrased Questions}

During validation, 39 rephrased question templates were flagged by a majority of annotators as 'BAD' due to semantic misalignment or lack of clarity. These were manually corrected. Table~\ref{tab:appendix_rephrase_examples} presents examples of original templates, their initially generated (problematic) rephrasings, and the manually corrected versions used in the final dataset.

\begin{table*}[!ht] 
\small
\centering
\renewcommand{\arraystretch}{1.2} 
\setlength{\tabcolsep}{4pt}
\begin{tabularx}{\textwidth}{@{} l p{0.28\textwidth} p{0.28\textwidth} p{0.28\textwidth} @{}} 
\toprule
\textbf{ID} & \textbf{Original Template} & \textbf{Initial (BAD) Rephrasing} & \textbf{Corrected (GOOD) Rephrasing} \\
\midrule
Ji-ko-25 & What are the family-related holidays in \{country\}? & \textit{Template:} During which occasion is \{answer\} celebrated? \newline \textit{Placeholder:} this celebration & \textit{Template:} In which country/region is \{answer\} the holiday most associated with family? \newline \textit{Placeholder:} this holiday \\
\addlinespace 
Sa-en-22 & What is the most famous university in \{country\} known for its sports team? & \textit{Template:} At which university is \{answer\} known for its sports team? \newline \textit{Placeholder:} this university & \textit{Template:} \{answer\} is the most famous university known for its sports team in which country/region? \newline \textit{Placeholder:} this university \\
\addlinespace
New-as-01 & What is the most famous religious holiday in \{country\}? & \textit{Template:} During which occasion is \{answer\} celebrated? \newline \textit{Placeholder:} this religious holiday & \textit{Template:} In which country/region is \{answer\} the most famous religious holiday? \newline \textit{Placeholder:} this holiday \\
\bottomrule
\end{tabularx}
\vspace{-1em} 
\caption{Examples of Manually Corrected Rephrased Question Templates.}
\label{tab:appendix_rephrase_examples}
\end{table*}

The initial rephrasings often failed to capture the specific nuance of the original question (e.g., focusing only on celebration rather than family association or religious nature) or created grammatically awkward structures. The manual corrections aimed to restore the original intent while adhering to the required (Entity $\rightarrow$ Region) format.

\subsection{Examples of 'BAD' Images}

\begin{figure}[!ht]
    \centering
    \includegraphics[width=0.45\textwidth]{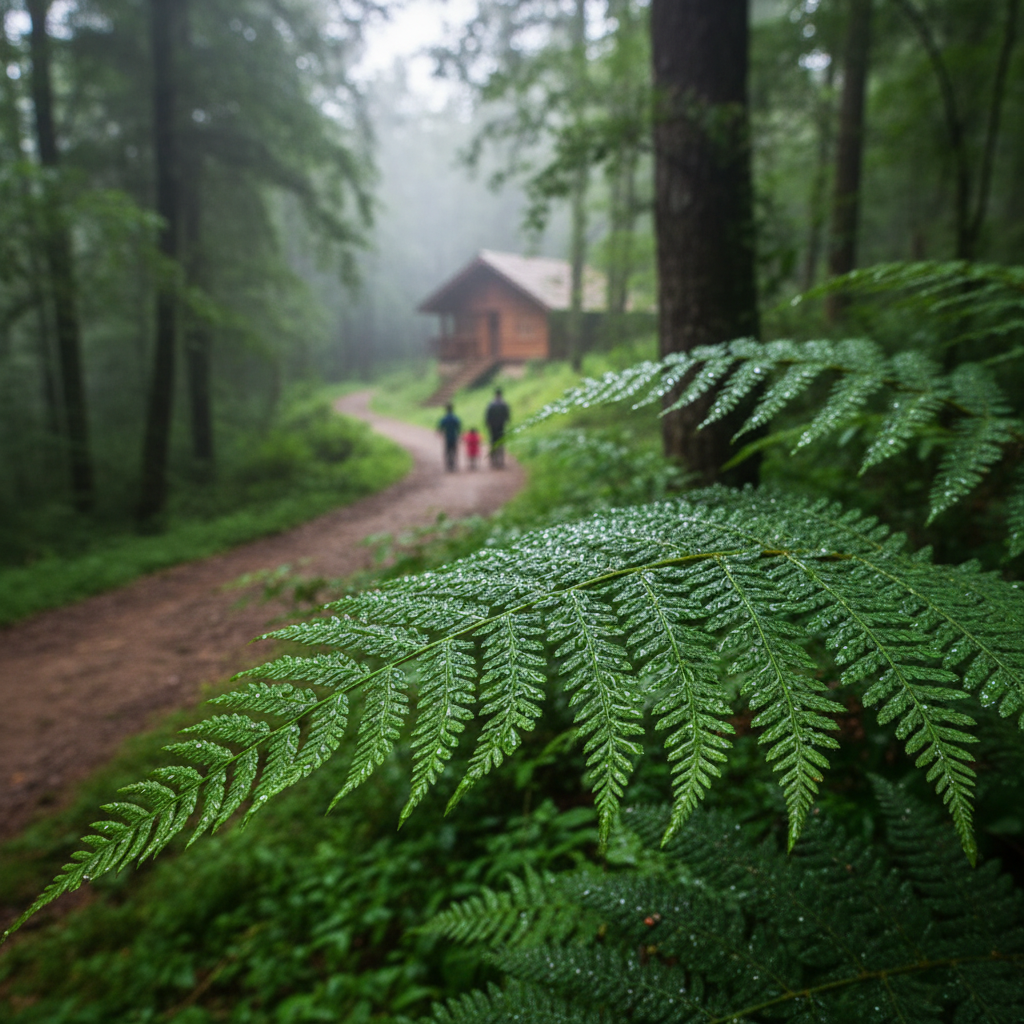}
    \caption[Example of 'BAD' Image (Ambiguous Representation)]{
        \textbf{ID:} Ca-sp-45, \textbf{Topic:} Family, \textbf{Region:} Iran, \textbf{Target Answer:} `north'. \\
        \textbf{Reason Flagged 'BAD':} The image of a family in a forest is too generic and lacks specific visual cues to represent a destination in the 'north' of Iran, failing to convey the intended concept. Failure mode: \textit{Ambiguous/Unclear Representation}.
    }
    \label{fig:bad_image_example1}
\end{figure}

\begin{figure}[!ht]
    \centering
    \includegraphics[width=0.45\textwidth]{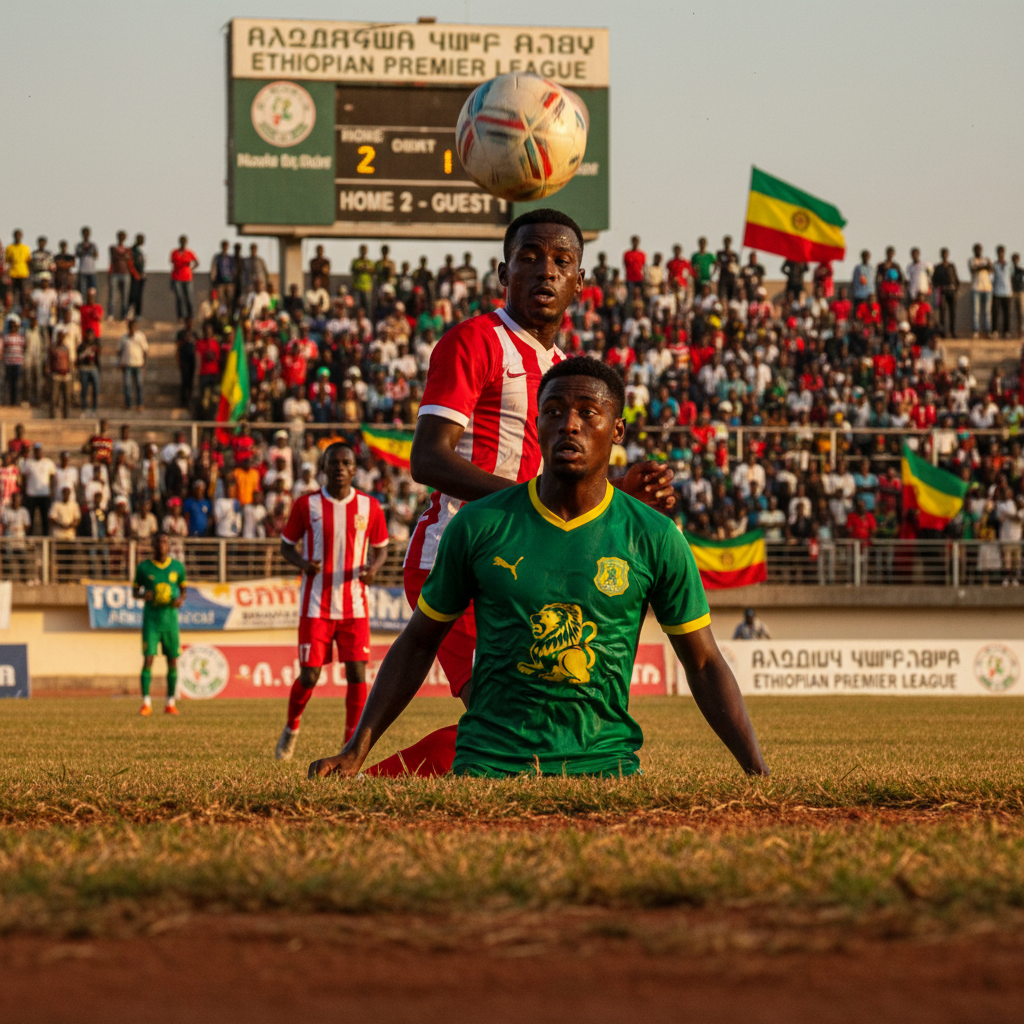}
    \caption[Example of 'BAD' Image (Severe Artefacts)]{
        \textbf{ID:} Th-en-03, \textbf{Topic:} Sport, \textbf{Region:} Ethiopia, \textbf{Target Answer:} `football league'. \\
        \textbf{Reason Flagged 'BAD':} The image contains severe generation artefacts, such as the player's lower body being clipped by the ground, which makes it look unnatural and distorted. Failure mode: \textit{Unusably Distorted}.
    }
    \label{fig:bad_image_example2}
\end{figure}

\begin{figure}[!ht]
    \centering
    \includegraphics[width=0.45\textwidth]{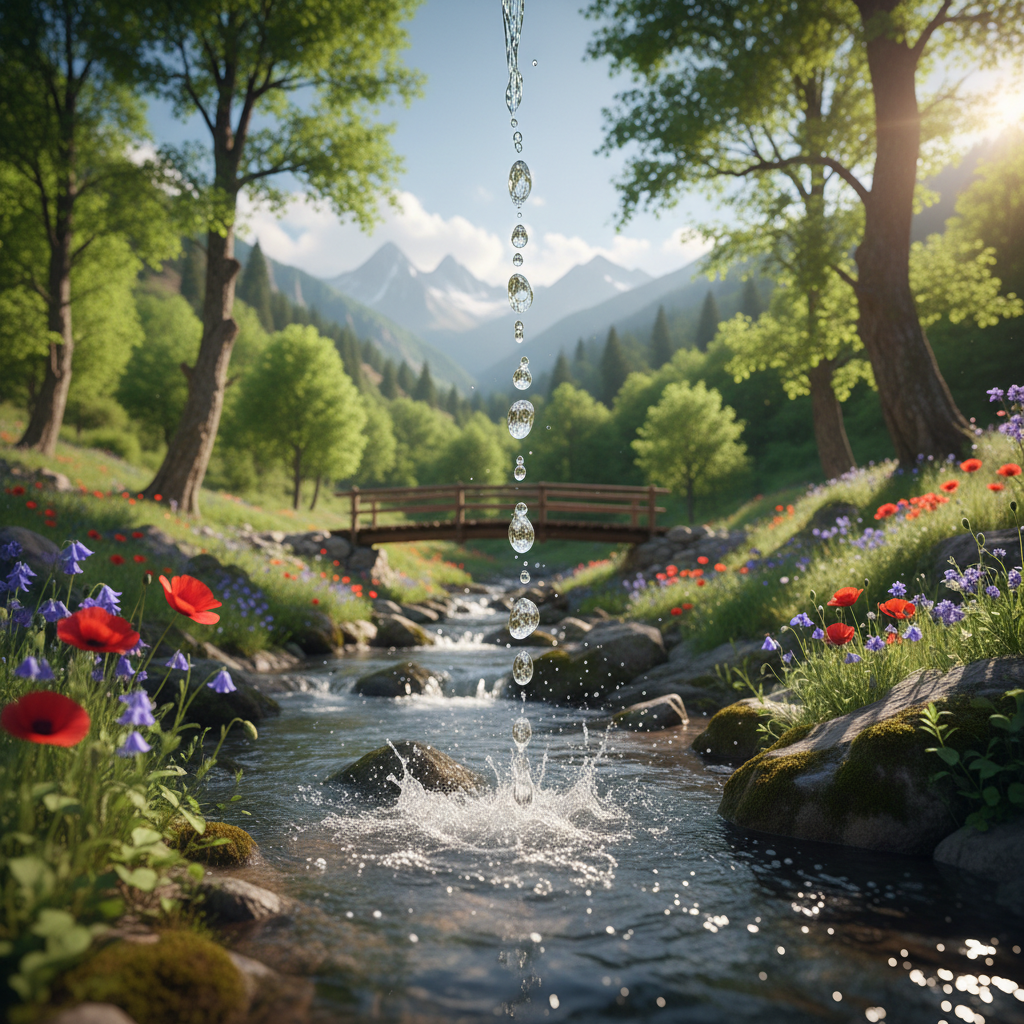}
    \caption[Example of 'BAD' Image (Inaccurate Representation)]{
        \textbf{ID:} Na-ko-45, \textbf{Topic:} Holidays/Celebration/Leisure, \textbf{Region:} Azerbaijan, \textbf{Target Answer:} `qabala'. \\
        \textbf{Reason Flagged 'BAD':} The generated landscape is too generic and does not accurately reflect the visual characteristics of the actual destination, Qabala in Azerbaijan. Failure mode: \textit{Inaccurate Regional Representation}.
    }
    \label{fig:bad_image_example3}
\end{figure}

Validation also identified images that failed to accurately or appropriately represent the target concept. Figures \ref{fig:bad_image_example1} through \ref{fig:bad_image_example3} illustrate examples flagged as 'BAD' by all three annotators, showcasing common failure modes such as ambiguous representation, severe generation artefacts, and a lack of regional specificity.

These examples highlight ongoing challenges in automated image generation, particularly in creating images that are not only free of artefacts but also visually specific and culturally/geographically accurate. The validation step was crucial for identifying such failures.

\newpage

\section{Prompts Used in Dataset Curation and Evaluation}
\label{appendix:prompts}

This section details the prompts provided to Large Language Models (LLMs) and image generation models during the automated stages of the \textbf{\texttt{BLEnD-Vis}} dataset construction pipeline.

\subsection{Tangibility Classification Prompt (GPT-4o)}
\textbf{Purpose:} Classify if a question template and its answers refer to tangible concepts suitable for image generation (Figure~\ref{fig:tangibility_prompt}). Used in Step 2.

\begin{figure}[!t]
\centering
\begin{tcolorbox}[colback=white!95!black, colframe=black!75] 
\scriptsize 
\textbf{User Prompt:}\\
\texttt{Please determine if the following question and its answers across different regions can be visually represented in an image.} \\
\texttt{} \\
\texttt{Question: \textcolor{violet}{\{question\_template\}}} \\
\texttt{} \\
\texttt{Answers across regions:} \\
\texttt{\textcolor{teal}{\{formatted\_answers\}}} \\ 
\texttt{} \\
\texttt{Task: Analyse whether this question and its answers reference tangible concepts that can be clearly depicted in an image. Also analyse whether answers reference specific entities (people or place).} \\
\texttt{} \\
\texttt{Classification criteria examples:} \\
\texttt{- Tangible (include): food, drinks, items, sports, occupations, festivals, commodities, famous people, infrastructure, religious symbols, instruments, clothes, animals, common physical activities} \\
\texttt{- Intangible (exclude): dates/times, ages, musical genres, languages, software, numbers, insurance, abstract concepts, entrance exams, education subjects} \\
\texttt{} \\
\texttt{Also label whether any answer references a specific person or place.} \\
\texttt{} \\
\texttt{For consistency:} \\
\texttt{1. A question about "What is the most popular X" can be tangible if X itself can be visually depicted} \\
\texttt{2. Questions about specific quantities (e.g., "How many hours...") are generally intangible} \\
\texttt{3. Questions about time periods, ages, or numeric data are intangible} \\
\texttt{4. Consider the question AND the answers - all must be visually representable} \\
\texttt{} \\
\texttt{Please respond with:} \\
\texttt{- is\_tangible: true/false} \\
\texttt{- reason: brief explanation for your decision} \\
\texttt{- specific\_entity: true/false (whether any answer references a specific person or place)} \\
\texttt{} \\
\texttt{Format your response in JSON:} \\
\texttt{\{} \\
\texttt{  "is\_tangible": boolean,} \\
\texttt{  "reason": "string",} \\
\texttt{  "specific\_entity": boolean} \\
\texttt{\}} \\
\end{tcolorbox}
\caption{Prompt used for Tangibility Filtering (Step 2).}
\label{fig:tangibility_prompt}
\end{figure}

\subsection{Question Rephrasing \& Placeholder Generation Prompt (GPT-4o)}
\textbf{Purpose:} Rephrase the original question template to invert the query (Entity $\rightarrow$ Region) and generate a generic placeholder text for the VQA-style format (Figure~\ref{fig:rephrasing_prompt}). Used in Step 3.

\begin{figure}[!t]
\centering
\begin{tcolorbox}[colback=white!95!black, colframe=black!75] 
\scriptsize 
\textbf{User Prompt:}\\
\texttt{Your task is to rephrase the original question template so that the concept represented by the original answers becomes the new subject. The goal is to create a single, natural-sounding rephrased question template (using an \{answer\} placeholder) and a single generic image placeholder text. This rephrased template and placeholder should work well regardless of which specific answer (from the examples below or similar ones) is eventually used. The rephrased question should implicitly ask for the country/region where the \{answer\} fits the description, and it must NOT contain any country/region information itself.} \\
\texttt{} \\
\texttt{Original question template:} \\
\texttt{'\textcolor{violet}{\{question\_template\}}'} \\
\texttt{} \\
\texttt{This template can produce questions for different countries, with answers such as: \textcolor{teal}{\{sample\_answers\_str\}}.} \\ 
\texttt{} \\
\texttt{Based on the original template and the type of answers it elicits, please generate:} \\
\texttt{1. A rephrased question template. It MUST include the exact placeholder \{answer\}. It should be phrased to ask 'In which country/region...?' or similar, making the location the expected answer.} \\
\texttt{2. A short, generic text (under 10 words) to be used as a placeholder when an image of the answer is shown alongside the rephrased question template. This text should describe the *type* of thing the answer is (e.g., 'this food', 'this item', 'this tradition', 'this landmark', 'this event').} \\
\texttt{} \\
\texttt{Separate the rephrased template and the image placeholder text with the special sequence: [IMAGE\_PLACEHOLDER\_SEP]} \\
\texttt{} \\
\texttt{Do not add explanations. Ensure the \{answer\} placeholder is present in the rephrased template.} \\
\texttt{} \\
\texttt{Example 1:} \\
\texttt{Original question template: 'What is a popular street food snack in \{country\}?'} \\
\texttt{Sample answers: 'churros', 'takoyaki', 'pretzels'} \\
\texttt{Your response should be:} \\
\texttt{In which country/region is \{answer\} a popular street food snack?} \\
\texttt{[IMAGE\_PLACEHOLDER\_SEP]} \\
\texttt{this street food} \\
\texttt{} \\
\texttt{Example 2:} \\
\texttt{Original question template: 'What is the name of the traditional clothing worn by women in \{country\}?'} \\
\texttt{Sample answers: 'kimono', 'sari', 'dirndl'} \\
\texttt{Your response should be:} \\
\texttt{In which country/region is \{answer\} the traditional clothing worn by women?} \\
\texttt{[IMAGE\_PLACEHOLDER\_SEP]} \\
\texttt{this traditional clothing} \\
\texttt{} \\
\texttt{Example 3:} \\
\texttt{Original question template: 'What is a common way to greet someone in \{country\}?'} \\
\texttt{Sample answers: 'bowing', 'handshake', 'wai'} \\
\texttt{Your response should be:} \\
\texttt{In which country/region is \{answer\} a common way to greet someone?} \\
\texttt{[IMAGE\_PLACEHOLDER\_SEP]} \\
\texttt{this greeting} \\
\texttt{} \\
\texttt{Now, based on the provided original question template and answer examples, generate the rephrased template and image placeholder.} \\
\texttt{} \\
\texttt{Rephrased question template and image placeholder:}
\end{tcolorbox}
\caption{Prompt used for Question Rephrasing \& Placeholder Generation (Step 3).}
\label{fig:rephrasing_prompt}
\end{figure}

\subsection{Image Generation Prompt (Gemini-Flash-2.5)}
\textbf{Purpose:} Generate a culturally contextualised image representing a specific answer entity, using the original question for context (Figure~\ref{fig:image_gen_prompt}). Used in Step 4.

\begin{figure}[!t]
\centering
\begin{tcolorbox}[colback=white!95!black, colframe=black!75] 
\scriptsize 
\textbf{User Prompt:}\\
\texttt{\textcolor{violet}{\{question\_template\_instance\}}} \\ 
\texttt{} \\
\texttt{Given the above question, generate a photorealistic and natural-looking photo of the following:} \\
\texttt{photo of \textcolor{OliveGreen}{\{original\_answer\}} in \textcolor{blue}{\{country\}}. 4k HDR beautiful, natural, Macro, High detail, focus, full-color.}
\end{tcolorbox}
\caption{Prompt used for Image Generation (Step 4).}
\label{fig:image_gen_prompt}
\end{figure}

\subsection{VLM Evaluation Prompt Template}
\textbf{Purpose:} General template used to query Vision-Language Models for all three MCQ formats (Original Text, Rephrased Text, VQA-Style) in \textbf{\texttt{BLEnD-Vis}}. For VQA-Style, an image is provided to the model preceding this textual prompt (Figure~\ref{fig:evaluation_prompt_template}).

\begin{figure}[!t]
\centering
\begin{tcolorbox}[colback=white!95!black, colframe=black!75]
\scriptsize
\textbf{User Prompt:}\\
\texttt{\textcolor{purple}{\{formatted\_question\_text\}}} \\ 
\texttt{Without any explanation, choose only one from the given alphabet choices(e.g., A, B, C). Provide as JSON format: \{\{"answer\_choice":""\}\}} \\
\texttt{} \\
\texttt{A. \textcolor{orange}{\{choice\_A\_text\}}} \\
\texttt{B. \textcolor{orange}{\{choice\_B\_text\}}} \\
\texttt{C. \textcolor{orange}{\{choice\_C\_text\}}} \\
\texttt{D. \textcolor{orange}{\{choice\_D\_text\}}} \\
\texttt{} \\
\texttt{Answer:}
\end{tcolorbox}
\caption{General prompt template used for VLM evaluation across all \textbf{\texttt{BLEnD-Vis}} MCQ formats. For VQA, an image precedes this text.}
\label{fig:evaluation_prompt_template}
\end{figure}
\newpage

\section{Analysis of Cross-Modal Agreement Patterns}
\label{appendix:agreement_analysis}
To provide deeper insight into model behavior, we analysed the patterns of agreement and disagreement between the Rephrased text-only ($R$) and VQA ($V$) formats across all models. We categorised each instance into one of five outcomes. The quantitative breakdowns by topic and region are presented, followed by qualitative examples illustrating each pattern.

\subsection{Quantitative Analysis of Agreement Patterns}

Tables \ref{tab:appendix_agreement_topic} and \ref{tab:appendix_agreement_region} show the distribution of outcomes. The analysis reveals two key patterns:

\textbf{Systematic Bias in Low-Resource Regions:} Models often share the same incorrect answer across both text and vision for low-resource regions, suggesting entrenched biases. The `Agree \& Incorrect` rate for regions like \textbf{North Korea (8.13\%)} and \textbf{Assam (6.87\%)} is over \textbf{3 times higher} than for the \textbf{US (2.00\%)}. This indicates that when models are uncertain, they converge on the same plausible (but wrong) high-resource answer in both modalities.
    
\textbf{Topic-Specific Modality Strengths:} In the topic breakdown, \textbf{Work life} has the highest `Agree \& Correct` rate (\textbf{51.71\%}), showing strong cross-modal understanding. However, it also has the highest `Disagree (R\_Correct)` rate (\textbf{14.33\%}). This suggests that concepts related to "Work life" (e.g., specific job roles or workplace norms) are well-represented textually but can be visually ambiguous or difficult to depict in a single image, causing the VQA modality to fail more often.

\newcommand{\stackcell}[1]{\begin{tabular}{@{}c@{}}#1\end{tabular}}

\begin{table}[!ht]
\small \centering
\renewcommand{\arraystretch}{0.9}
\setlength{\tabcolsep}{2pt}
\begin{tabular}{@{} l S[table-format=2.2] S[table-format=2.2] S[table-format=2.2] S[table-format=2.2] S[table-format=2.2] @{}}
\toprule
\textbf{Topic} & {\stackcell{Agree- \\ Corr}} & {\stackcell{Agree- \\ Incorr}} & {\stackcell{Disagree \\ (R\checkmark)}} & {\stackcell{Disagree \\ (V\checkmark)}} & {\stackcell{Disagree- \\ Incorr}} \\
\midrule
Food                         & 40.87 & 5.33 &  9.40 & 26.27 & 18.13 \\
Sport                        & 42.56 & 3.97 &  9.48 & 28.71 & 15.27 \\
Hols./Celeb. & 44.98 & 4.09 & 10.20 & 26.40 & 14.34 \\
Family                       & 35.98 & 4.74 &  8.33 & 33.90 & 17.06 \\
Work life                    & 51.71 & 2.96 & 14.33 & 19.44 & 11.55 \\
Education                    & 36.30 & 4.10 &  8.27 & 34.36 & 16.97 \\
\bottomrule
\end{tabular}
\caption{Cross-Modal Outcome Patterns by Topic (\%).}
\label{tab:appendix_agreement_topic}
\end{table}

\begin{table}[!ht]
\small \centering
\renewcommand{\arraystretch}{0.8}
\setlength{\tabcolsep}{1.5pt}
\begin{tabular}{@{} l S[table-format=2.2] S[table-format=2.2] S[table-format=2.2] S[table-format=2.2] S[table-format=2.2] @{}}
\toprule
\textbf{Region} & {\stackcell{Agree- \\ Corr}} & {\stackcell{Agree- \\ Incorr}} & {\stackcell{Disagree \\ (R\checkmark)}} & {\stackcell{Disagree \\ (V\checkmark)}} & {\stackcell{Disagree- \\ Incorr}} \\
\midrule
Assam            & 27.99 & 6.87 &  9.98 & 29.09 & 26.08 \\
South Korea      & 56.07 & 2.54 & 10.92 & 22.11 &  8.37 \\
Mexico           & 46.57 & 3.64 &  8.43 & 30.54 & 10.82 \\
China            & 59.42 & 2.41 &  8.34 & 22.78 &  7.05 \\
Indonesia        & 48.30 & 3.60 & 10.71 & 25.18 & 12.22 \\
Ethiopia         & 29.28 & 4.82 &  8.18 & 35.40 & 22.33 \\
Greece           & 40.84 & 4.49 &  7.57 & 33.73 & 13.37 \\
Spain            & 45.79 & 3.99 &  9.42 & 27.06 & 13.75 \\
Iran             & 35.93 & 4.37 & 10.35 & 34.34 & 15.01 \\
US               & 66.34 & 2.00 & 11.09 & 14.53 &  6.04 \\
North Korea      & 25.08 & 8.13 & 10.41 & 30.38 & 26.00 \\
UK               & 60.63 & 2.57 &  9.31 & 20.52 &  6.96 \\
West Java        & 33.96 & 6.37 & 10.33 & 25.31 & 24.03 \\
Azerbaijan       & 33.44 & 4.34 & 10.03 & 32.12 & 20.07 \\
Algeria          & 25.38 & 6.64 & 11.38 & 27.96 & 28.64 \\
North. Nigeria & 35.93 & 3.92 & 12.40 & 31.17 & 16.58 \\
\bottomrule
\end{tabular}
\caption{Cross-Modal Outcome Patterns by Region (\%).}
\label{tab:appendix_agreement_region}
\end{table}

\subsection{Qualitative Examples of Agreement Patterns}
We present five examples from the Qwen2.5-VL-32B model to illustrate each cross-modal outcome. Each example shows the model's response to both text-only (Rephrased) and image-based (VQA) versions of the same question.

\definecolor{correctgreen}{HTML}{006600}
\definecolor{incorrectred}{HTML}{B00020}

\begin{table}[!ht]
\small
\centering
\begin{tabular}{p{0.28\linewidth} | p{0.6\linewidth}}
\toprule
\textbf{MCQ Info \& Choices} & \textbf{Model Responses (Qwen2.5-VL-32B)} \\
\midrule
\textbf{MCQID:} Th-en-15\_18 \newline
\textbf{Topic:} Sport \newline
\textbf{Entity:} hockey \newline
\textbf{Choices:} \newline A. Assam, \newline \textbf{B. US (Correct)}, \newline C. UK, \newline D. Ethiopia
& 
\textbf{\textit{Rephrased (Text-Only):}} In which country/region is hockey the most popular winter sport? \newline 
\textbf{Answer:} \textcolor{correctgreen}{B. US (Correct)} \newline\newline
\textbf{\textit{VQA (Image):}} In which country/region is this winter sport the most popular winter sport? \newline
\includegraphics[height=2cm]{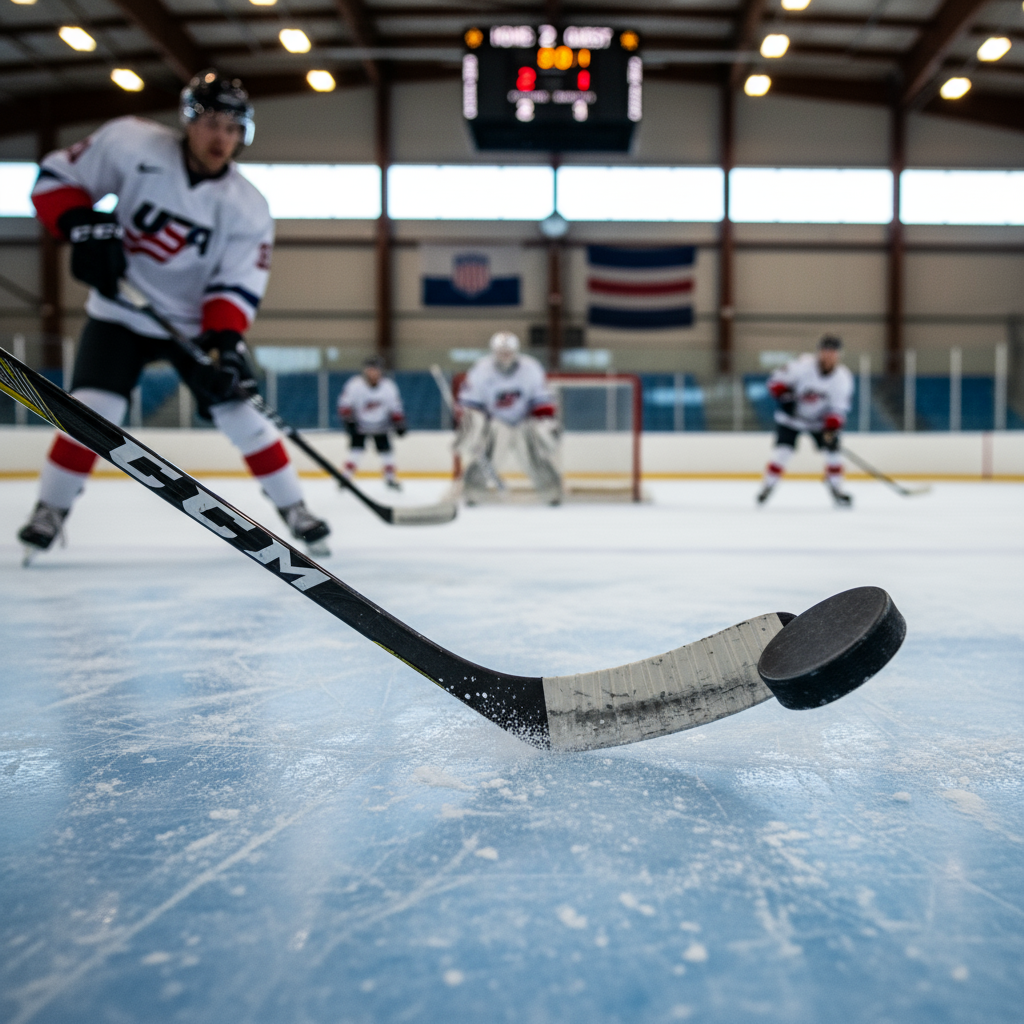} \newline
\textbf{Answer:} \textcolor{correctgreen}{B. US (Correct)} \\
\bottomrule
\end{tabular}
\caption{Agree \& Correct -- The model correctly answers in both modalities, demonstrating robust, well-grounded knowledge.}
\end{table}

\begin{table}[!ht]
\small
\centering
\begin{tabular}{p{0.28\linewidth} | p{0.6\linewidth}}
\toprule
\textbf{MCQ Info \& Choices} & \textbf{Model Responses (Qwen2.5-VL-32B)} \\
\midrule
\textbf{MCQID:} Jo-sp-02\_65 \newline
\textbf{Topic:} Sport \newline
\textbf{Entity:} chess \newline
\textbf{Choices:} \newline A. Mexico, \newline B. Northern Nigeria, \newline \textbf{C. Greece (Correct)}, \newline D. North Korea
& 
\textbf{\textit{Rephrased (Text-Only):}} In which country/region is chess the most popular sport played without a ball? \newline
\textbf{Answer:} \textcolor{incorrectred}{D. North Korea (Incorrect)} \newline\newline
\textbf{\textit{VQA (Image):}} In which country/region is this sport the most popular sport played without a ball? \newline
\includegraphics[height=2cm]{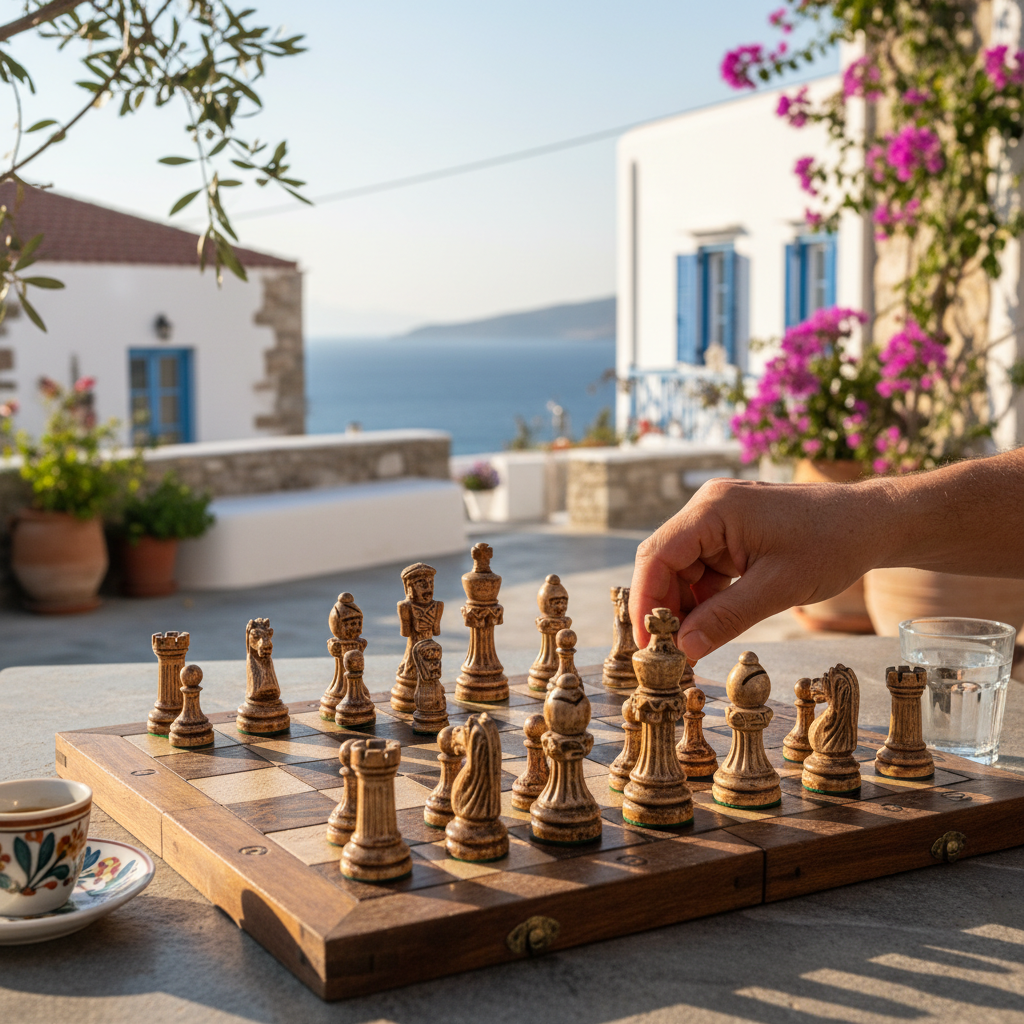} \newline
\textbf{Answer:} \textcolor{correctgreen}{C. Greece (Correct)} \\
\bottomrule
\end{tabular}
\caption{Disagree (VQA Corrects Text) -- The model fails textually but the visual cue helps it recover the correct answer, highlighting the value of grounding.}
\end{table}

\begin{table}[!ht]
\small
\centering
\begin{tabular}{p{0.28\linewidth} | p{0.6\linewidth}}
\toprule
\textbf{MCQ Info \& Choices} & \textbf{Model Responses (Qwen2.5-VL-32B)} \\
\midrule
\textbf{MCQID:} New-en-59\_13 \newline
\textbf{Topic:} Family \newline
\textbf{Entity:} russia \newline
\textbf{Choices:} \newline A. UK, \newline B. Algeria, \newline \textbf{C. Azerbaijan (Correct)}, \newline D. South Korea
& 
\textbf{\textit{Rephrased (Text-Only):}} In which country/region is russia the most popular destination for families to emigrate to? \newline
\textbf{Answer:} \textcolor{correctgreen}{C. Azerbaijan (Correct)} \newline\newline
\textbf{\textit{VQA (Image):}} In which country/region is this destination the most popular destination for families to emigrate to? \newline
\includegraphics[height=2cm]{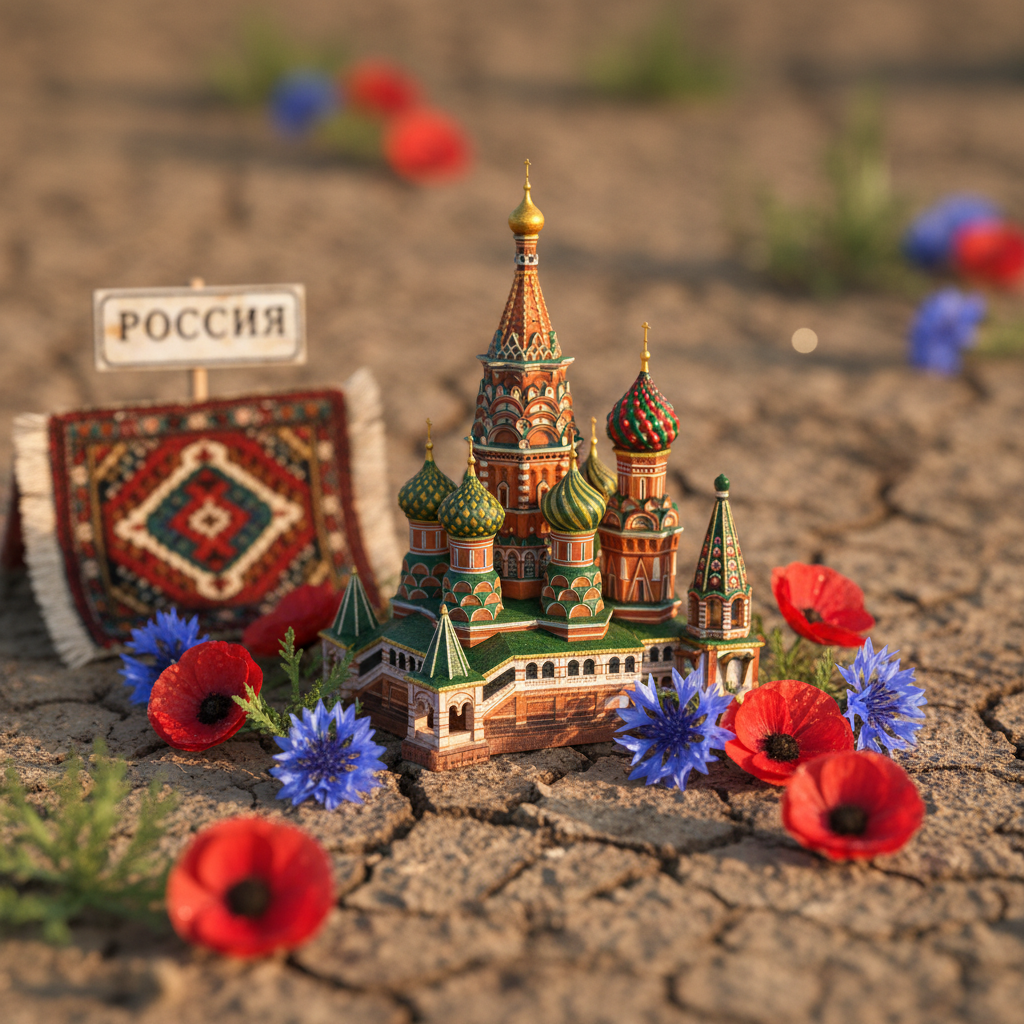} \newline
\textbf{Answer:} \textcolor{incorrectred}{A. UK (Incorrect)} \\
\bottomrule
\end{tabular}
\caption{Disagree (Text Corrects VQA) -- The model succeeds textually but fails with the visual input.}
\end{table}

\begin{table}[!ht]
\small
\centering
\begin{tabular}{p{0.28\linewidth} | p{0.6\linewidth}}
\toprule
\textbf{MCQ Info \& Choices} & \textbf{Model Responses (Qwen2.5-VL-32B)} \\
\midrule
\textbf{MCQID:} Th-en-15\_1 \newline
\textbf{Topic:} Sport \newline
\textbf{Entity:} skiing \newline
\textbf{Choices:} \newline \textbf{A. UK (Correct)}, \newline B. Assam, \newline C. US, \newline D. North Korea
& 
\textbf{\textit{Rephrased (Text-Only):}} In which country/region is skiing the most popular winter sport? \newline
\textbf{Answer:} \textcolor{incorrectred}{C. US (Incorrect)} \newline\newline
\textbf{\textit{VQA (Image):}} In which country/region is this winter sport the most popular winter sport? \newline
\includegraphics[height=2cm]{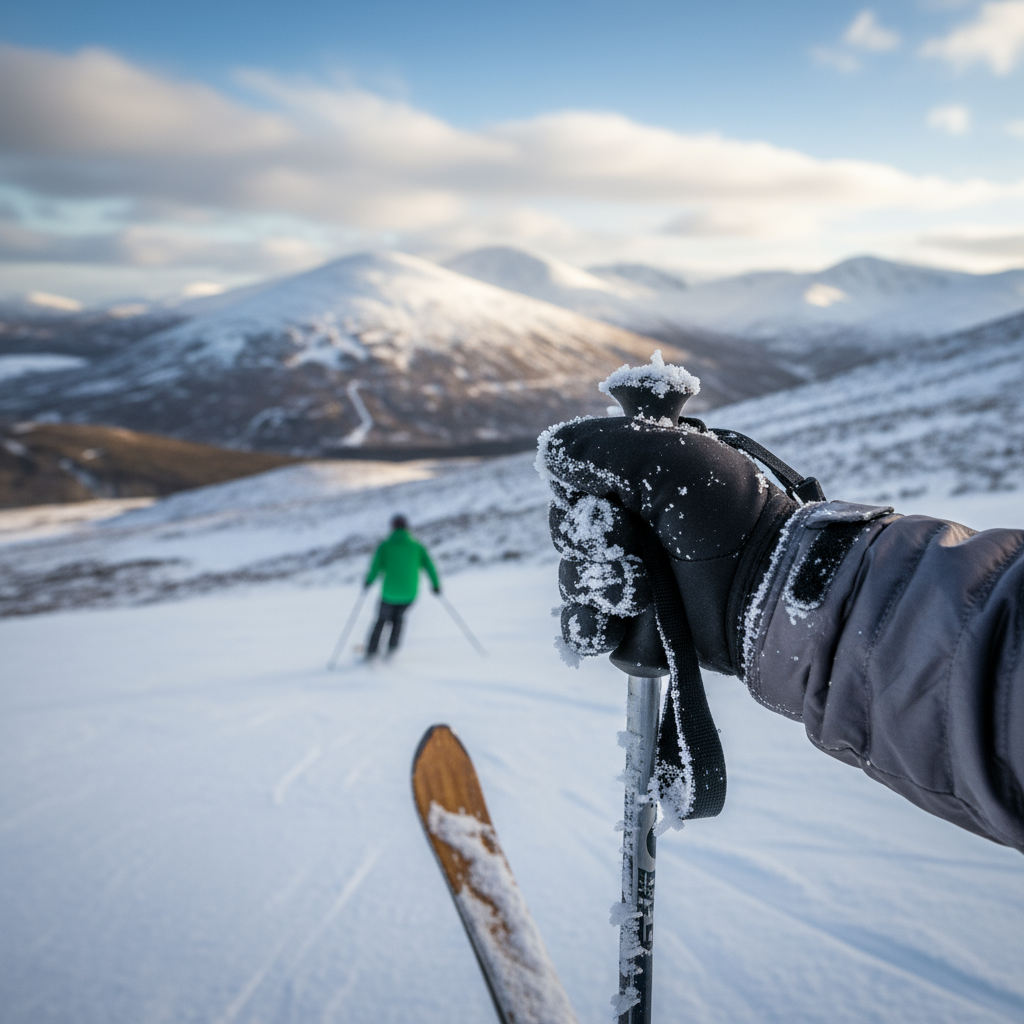} \newline
\textbf{Answer:} \textcolor{incorrectred}{C. US (Incorrect)} \\
\bottomrule
\end{tabular}
\caption{Agree \& Incorrect -- The model provides the same incorrect answer in both modalities, indicating a shared, systematic bias.}
\end{table}

\begin{table}[!ht]
\small
\centering
\begin{tabular}{p{0.28\linewidth} | p{0.6\linewidth}}
\toprule
\textbf{MCQ Info \& Choices} & \textbf{Model Responses (Qwen2.5-VL-32B)} \\
\midrule
\textbf{MCQID:} Jod-ch-15\_10 \newline
\textbf{Topic:} Food \newline
\textbf{Entity:} snail \newline
\textbf{Choices:} \newline A. South Korea, \newline B. US, \newline \textbf{C. Assam (Correct)}, \newline D. UK
& 
\textbf{\textit{Rephrased (Text-Only):}} In which country/region is snail a popular type of seafood? \newline
\textbf{Answer:} \textcolor{incorrectred}{D. UK (Incorrect)} \newline\newline
\textbf{\textit{VQA (Image):}} In which country/region is this seafood a popular type of seafood? \newline
\includegraphics[height=2cm]{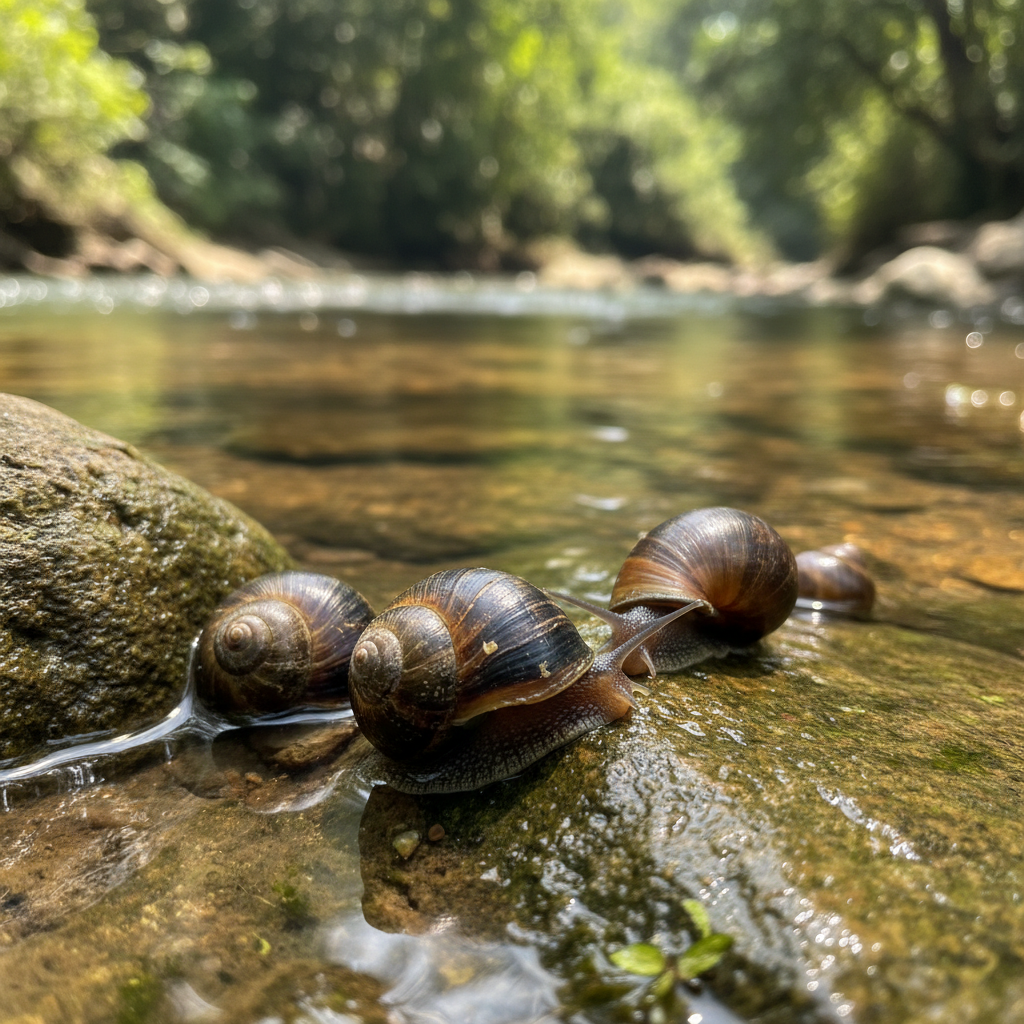} \newline
\textbf{Answer:} \textcolor{incorrectred}{A. South Korea (Incorrect)} \\
\bottomrule
\end{tabular}
\caption{Disagree \& Both Incorrect -- The model is incorrect in both modalities and provides different answers, suggesting a lack of knowledge.}
\end{table}

\onecolumn

\definecolor{correctgreen}{HTML}{006600} 

\newcolumntype{P}[1]{>{\raggedright\arraybackslash}p{#1}}

\section{Examples of Parallel MCQ Formats}
\label{appendix:mcq_examples}

\begin{table*}[!ht]
\small
\centering
\renewcommand{\arraystretch}{0.8} 
\setlength{\tabcolsep}{6pt}

\begin{tabular}{@{} l l P{0.6\textwidth} @{}}
\toprule
\textbf{MCQ-ID} & \textbf{MCQ Format} & \textbf{Question \& Options} \\
\midrule

\multirow{3}{*}{\textbf{Al-en-01\_1}} 
 & Original (Region $\rightarrow$ Entity) &
    \textbf{Q:} What is a common snack for preschool kids in West Java? \newline 
    \textbf{Options:} \newline
    A. toast \newline
    B. candy \newline
    C. mashed potato rice \newline
    \textcolor{correctgreen}{D. jelly} \\
\cmidrule(l){2-3} 
 & Rephrased (Entity $\rightarrow$ Region) &
    \textbf{Q:} For which country/region is jelly a common snack for preschool kids? \newline 
    \textbf{Options:} \newline 
    A. Greece \newline
    B. North Korea \newline
    C. Assam \newline
    \textcolor{correctgreen}{D. West Java} \\
\cmidrule(l){2-3}
 & VQA-Style (Image $\rightarrow$ Region) &
    \textbf{Image:} \newline
    \mbox{\includegraphics[height=2.5cm]{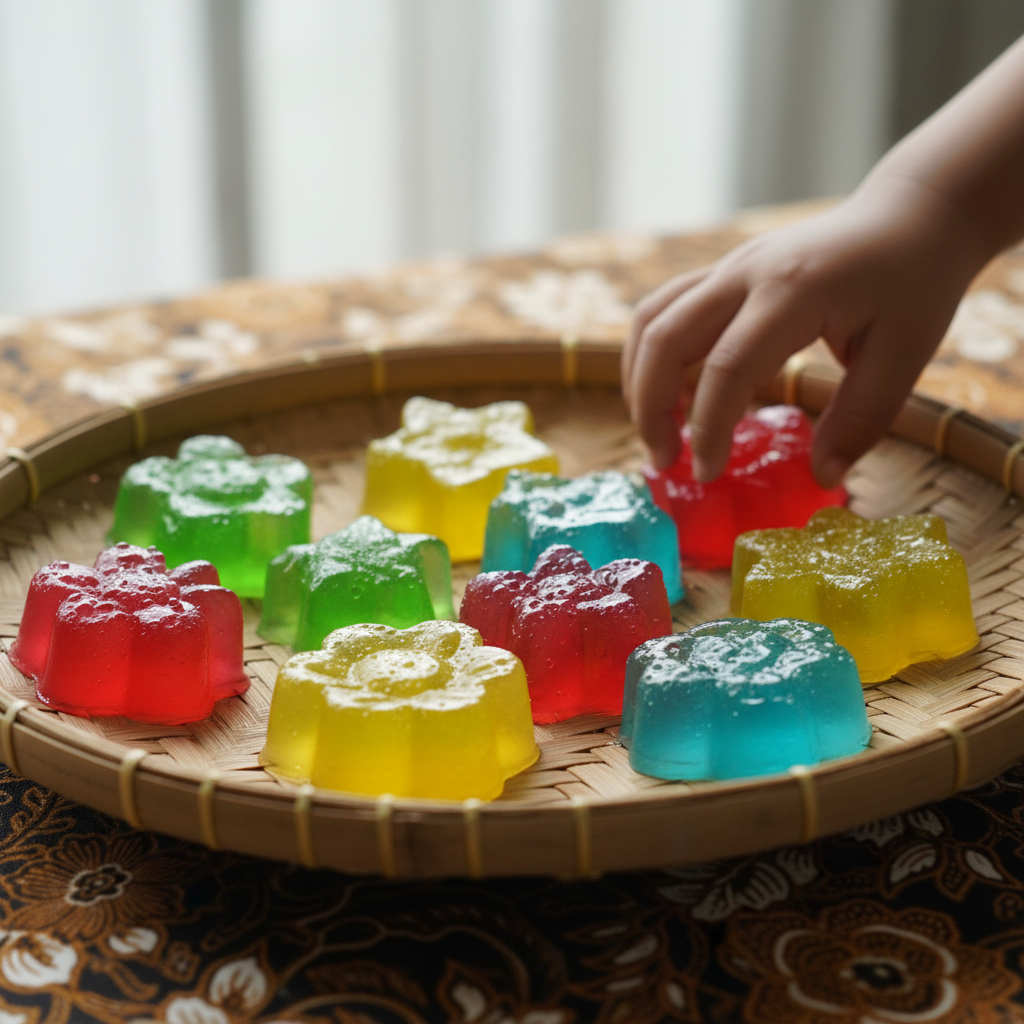}} \quad 
    \newline
    \textbf{Q:} For which country/region is this snack a common snack for preschool kids? \newline 
    \textbf{Options:} \newline
    A. Greece \newline
    B. North Korea \newline
    C. Assam \newline
    \textcolor{correctgreen}{D. West Java} \\

\midrule 

\multirow{3}{*}{\textbf{Th-en-01\_9}} 
 & Original (Region $\rightarrow$ Entity) &
    \textbf{Q:} What is the most popular summer sport in Ethiopia? \newline 
    \textbf{Options:} \newline
    A. volleyball \newline
    \textcolor{correctgreen}{B. running} \newline
    C. swimming \newline
    D. badminton \\
\cmidrule(l){2-3}
 & Rephrased (Entity $\rightarrow$ Region) &
    \textbf{Q:} In which country/region is running the most popular summer sport? \newline 
    \textbf{Options:} \newline
    A. Spain \newline
    \textcolor{correctgreen}{B. Ethiopia} \newline
    C. Azerbaijan \newline
    D. Indonesia \\
\cmidrule(l){2-3}
 & VQA-Style (Image $\rightarrow$ Region) &
    \textbf{Image:} \newline
    \mbox{\includegraphics[height=2.5cm]{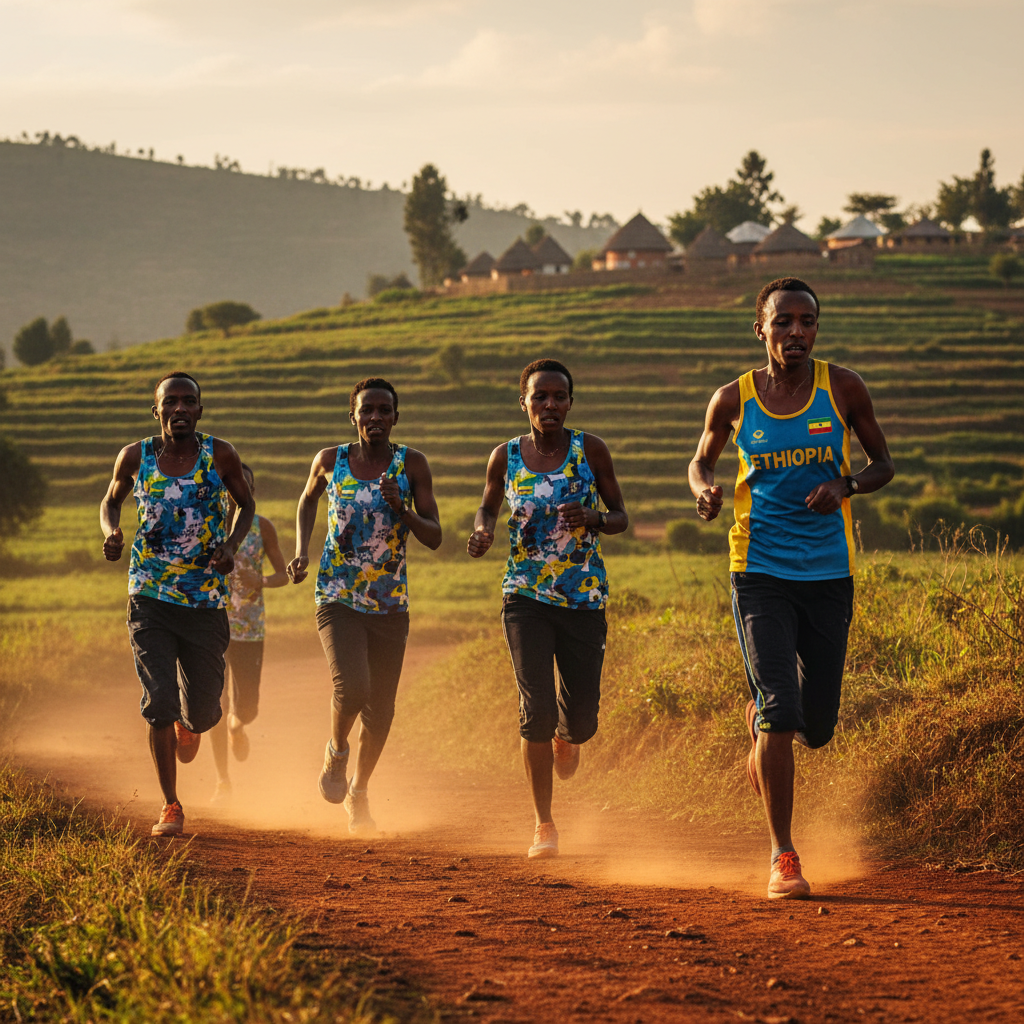}} \quad 
    \newline
    \textbf{Q:} In which country/region is this sport the most popular summer sport? \newline 
    \textbf{Options:} \newline
    A. Spain \newline
    \textcolor{correctgreen}{B. Ethiopia} \newline
    C. Azerbaijan \newline
    D. Indonesia \\

\bottomrule
\end{tabular}
\caption{Examples of Parallel MCQ Formats in \textbf{\texttt{BLEnD-Vis}}.}
\label{tab:parallel_mcq_examples_transposed}
\end{table*}

%% file: emnlp2023.bib
@inproceedings{adilazuardaMeasuringModelingCulture2024,
  title = {Towards Measuring and Modeling ``Culture'' in {{LLMs}}: A Survey},
  shorttitle = {Towards Measuring and Modeling ``Culture'' in {{LLMs}}},
  booktitle = {Proceedings of the 2024 {{Conference}} on {{Empirical Methods}} in {{Natural Language Processing}}},
  author = {Adilazuarda, Muhammad Farid and Mukherjee, Sagnik and Lavania, Pradhyumna and Singh, Siddhant Shivdutt and Aji, Alham Fikri and O'Neill, Jacki and Modi, Ashutosh and Choudhury, Monojit},
  editor = {{Al-Onaizan}, Yaser and Bansal, Mohit and Chen, Yun-Nung},
  year = {2024},
  month = nov,
  pages = {15763--15784},
  publisher = {Association for Computational Linguistics},
  address = {Miami, Florida, USA},
  doi = {10.18653/v1/2024.emnlp-main.882},
  urldate = {2025-05-01},
  langid = {english},
}

@inproceedings{alkhamissiInvestigatingCulturalAlignment2024,
  title = {Investigating Cultural Alignment of Large Language Models},
  booktitle = {Proceedings of the 62nd {{Annual Meeting}} of the {{Association}} for {{Computational Linguistics}} ({{Volume}} 1: {{Long Papers}})},
  author = {AlKhamissi, Badr and ElNokrashy, Muhammad and Alkhamissi, Mai and Diab, Mona},
  editor = {Ku, Lun-Wei and Martins, Andre and Srikumar, Vivek},
  year = {2024},
  month = aug,
  pages = {12404--12422},
  publisher = {Association for Computational Linguistics},
  address = {Bangkok, Thailand},
  urldate = {2024-08-13},
  langid = {english},
}

@inproceedings{bauerSocialCommonsenseExplanation2023,
  title = {Social Commonsense for Explanation and Cultural Bias Discovery},
  booktitle = {Proceedings of the 17th {{Conference}} of the {{European Chapter}} of the {{Association}} for {{Computational Linguistics}}},
  author = {Bauer, Lisa and Tischer, Hanna and Bansal, Mohit},
  editor = {Vlachos, Andreas and Augenstein, Isabelle},
  year = {2023},
  month = may,
  pages = {3745--3760},
  publisher = {Association for Computational Linguistics},
  address = {Dubrovnik, Croatia},
  doi = {10.18653/v1/2023.eacl-main.271},
  urldate = {2025-05-01},
  langid = {english},
}

@misc{dammuTheyAreUncultured2024,
  title = {"{{They}} Are Uncultured": {{Unveiling Covert Harms}} and {{Social Threats}} in {{LLM Generated Conversations}}},
  shorttitle = {"{{They}} Are Uncultured"},
  author = {Dammu, Preetam Prabhu Srikar and Jung, Hayoung and Singh, Anjali and Choudhury, Monojit and Mitra, Tanushree},
  year = {2024},
  month = may,
  number = {arXiv:2405.05378},
  eprint = {2405.05378},
  primaryclass = {cs},
  publisher = {arXiv},
  doi = {10.48550/arXiv.2405.05378},
  urldate = {2024-08-06},
  archiveprefix = {arXiv},
}

@inproceedings{liuAreMultilingualLLMs2024,
  title = {Are Multilingual {{LLMs}} Culturally-Diverse Reasoners? {{An}} Investigation into Multicultural Proverbs and Sayings},
  shorttitle = {Are Multilingual {{LLMs}} Culturally-Diverse Reasoners?},
  booktitle = {Proceedings of the 2024 {{Conference}} of the {{North American Chapter}} of the {{Association}} for {{Computational Linguistics}}: {{Human Language Technologies}} ({{Volume}} 1: {{Long Papers}})},
  author = {Liu, Chen and Koto, Fajri and Baldwin, Timothy and Gurevych, Iryna},
  editor = {Duh, Kevin and Gomez, Helena and Bethard, Steven},
  year = {2024},
  month = jun,
  pages = {2016--2039},
  publisher = {Association for Computational Linguistics},
  address = {Mexico City, Mexico},
  doi = {10.18653/v1/2024.naacl-long.112},
  urldate = {2024-11-03},
  langid = {english},
}

@inproceedings{NEURIPS2024_77f089cd,
  title = {{{CulturePark}}: {{Boosting}} Cross-Cultural Understanding in Large Language Models},
  booktitle = {Advances in Neural Information Processing Systems},
  author = {Li, Cheng and Teney, Damien and Yang, Linyi and Wen, Qingsong and Xie, Xing and Wang, Jindong},
  editor = {Globerson, A. and Mackey, L. and Belgrave, D. and Fan, A. and Paquet, U. and Tomczak, J. and Zhang, C.},
  year = {2024},
  volume = {37},
  pages = {65183--65216},
  publisher = {Curran Associates, Inc.}
}

@inproceedings{NEURIPS2024_8eb88844,
  title = {{{BLEnD}}: A Benchmark for Llms on Everyday Knowledge in Diverse Cultures and Languages},
  booktitle = {Advances in {{Neural Information Processing Systems}}},
  author = {Myung, Junho and Lee, Nayeon and Zhou, Yi and Jin, Jiho and Putri, Rifki Afina and Antypas, Dimosthenis and Borkakoty, Hsuvas and Kim, Eunsu and {Perez-Almendros}, Carla and Ayele, Abinew Ali and {Guti{\'e}rrez-Basulto}, V{\'{\i}}ctor and {Ib{\'a}{\~n}ez-Garc{\'{\i}}a}, Yazm{\'{\i}}n and Lee, Hwaran and Muhammad, Shamsuddeen Hassan and Park, Kiwoong and Rzayev, Anar Sabuhi and White, Nina and Yimam, Seid Muhie and Pilehvar, Mohammad Taher and Ousidhoum, Nedjma and {Camacho-Collados}, Jose and Oh, Alice},
  editor = {Globerson, A. and Mackey, L. and Belgrave, D. and Fan, A. and Paquet, U. and Tomczak, J. and Zhang, C.},
  year = {2024},
  volume = {37},
  pages = {78104--78146},
  publisher = {Curran Associates, Inc.},
  langid = {english}
}

@inproceedings{NEURIPS2024_9a16935b,
  title = {{{CultureLLM}}: Incorporating Cultural Differences into Large Language Models},
  booktitle = {Advances in {{Neural Information Processing Systems}}},
  author = {Li, Cheng and Chen, Mengzhuo and Wang, Jindong and Sitaram, Sunayana and Xie, Xing},
  editor = {Globerson, A. and Mackey, L. and Belgrave, D. and Fan, A. and Paquet, U. and Tomczak, J. and Zhang, C.},
  year = {2024},
  volume = {37},
  pages = {84799--84838},
  publisher = {Curran Associates, Inc.},
  langid = {english}
}

@misc{pawarSurveyCulturalAwareness2024,
  title = {Survey of Cultural Awareness in Language Models: Text and Beyond},
  shorttitle = {Survey of Cultural Awareness in Language Models},
  author = {Pawar, Siddhesh and Park, Junyeong and Jin, Jiho and Arora, Arnav and Myung, Junho and Yadav, Srishti and Haznitrama, Faiz Ghifari and Song, Inhwa and Oh, Alice and Augenstein, Isabelle},
  year = {2024},
  month = oct,
  number = {arXiv:2411.00860},
  eprint = {2411.00860},
  publisher = {arXiv},
  doi = {10.48550/arXiv.2411.00860},
  urldate = {2024-11-08},
  archiveprefix = {arXiv},
  langid = {english},
}

@article{taoCulturalBiasCultural2024,
  title = {Cultural Bias and Cultural Alignment of Large Language Models},
  author = {Tao, Yan and Viberg, Olga and Baker, Ryan S and Kizilcec, Ren{\'e} F},
  year = {2024},
  month = sep,
  journal = {PNAS Nexus},
  volume = {3},
  number = {9},
  pages = {pgae346},
  issn = {2752-6542},
  doi = {10.1093/pnasnexus/pgae346},
  urldate = {2025-05-01},
  langid = {english},
}

@inproceedings{urailertprasertSEAVQASoutheastAsian2024,
  title = {{{SEA-VQA}}: {{Southeast Asian Cultural Context Dataset For Visual Question Answering}}},
  shorttitle = {{{SEA-VQA}}},
  booktitle = {Proceedings of the 3rd {{Workshop}} on {{Advances}} in {{Language}} and {{Vision Research}} ({{ALVR}})},
  author = {Urailertprasert, Norawit and Limkonchotiwat, Peerat and Suwajanakorn, Supasorn and Nutanong, Sarana},
  editor = {Gu, Jing and Fu, Tsu-Jui (Ray) and Hudson, Drew and Celikyilmaz, Asli and Wang, William},
  year = {2024},
  month = aug,
  pages = {173--185},
  publisher = {Association for Computational Linguistics},
  address = {Bangkok, Thailand},
  doi = {10.18653/v1/2024.alvr-1.15},
  urldate = {2024-11-12},
}

@misc{vayaniAllLanguagesMatter2025,
  title = {All Languages Matter: Evaluating {{LMMs}} on Culturally Diverse 100 Languages},
  shorttitle = {All Languages Matter},
  author = {Vayani, Ashmal and Dissanayake, Dinura and Watawana, Hasindri and Ahsan, Noor and Sasikumar, Nevasini and Thawakar, Omkar and Ademtew, Henok Biadglign and Hmaiti, Yahya and Kumar, Amandeep and Kuckreja, Kartik and Maslych, Mykola and Ghallabi, Wafa Al and Mihaylov, Mihail and Qin, Chao and Shaker, Abdelrahman M. and Zhang, Mike and Ihsani, Mahardika Krisna and Esplana, Amiel and Gokani, Monil and Mirkin, Shachar and Singh, Harsh and Srivastava, Ashay and Hamerlik, Endre and Izzati, Fathinah Asma and Maani, Fadillah Adamsyah and Cavada, Sebastian and Chim, Jenny and Gupta, Rohit and Manjunath, Sanjay and Zhumakhanova, Kamila and Rabevohitra, Feno Heriniaina and Amirudin, Azril and Ridzuan, Muhammad and Kareem, Daniya and More, Ketan and Li, Kunyang and Shakya, Pramesh and Saad, Muhammad and Ghasemaghaei, Amirpouya and Djanibekov, Amirbek and Azizov, Dilshod and Jankovic, Branislava and Bhatia, Naman and Cabrera, Alvaro and {Obando-Ceron}, Johan and Otieno, Olympiah and Farestam, Fabian and Rabbani, Muztoba and Baliah, Sanoojan and Sanjeev, Santosh and Shtanchaev, Abduragim and Fatima, Maheen and Nguyen, Thao and Kareem, Amrin and Aremu, Toluwani and Xavier, Nathan and Bhatkal, Amit and Toyin, Hawau and Chadha, Aman and Cholakkal, Hisham and Anwer, Rao Muhammad and Felsberg, Michael and Laaksonen, Jorma and Solorio, Thamar and Choudhury, Monojit and Laptev, Ivan and Shah, Mubarak and Khan, Salman and Khan, Fahad},
  year = {2025},
  month = apr,
  number = {arXiv:2411.16508},
  eprint = {2411.16508},
  primaryclass = {cs},
  publisher = {arXiv},
  doi = {10.48550/arXiv.2411.16508},
  urldate = {2025-05-01},
  archiveprefix = {arXiv},
  langid = {english},
}

@inproceedings{wangCDEvalBenchmarkMeasuring2024,
  title = {{{CDEval}}: {{A Benchmark}} for {{Measuring}} the {{Cultural Dimensions}} of {{Large Language Models}}},
  shorttitle = {{{CDEval}}},
  booktitle = {Proceedings of the 2nd {{Workshop}} on {{Cross-Cultural Considerations}} in {{NLP}}},
  author = {Wang, Yuhang and Zhu, Yanxu and Kong, Chao and Wei, Shuyu and Yi, Xiaoyuan and Xie, Xing and Sang, Jitao},
  editor = {Prabhakaran, Vinodkumar and Dev, Sunipa and Benotti, Luciana and Hershcovich, Daniel and Cabello, Laura and Cao, Yong and Adebara, Ife and Zhou, Li},
  year = {2024},
  month = aug,
  pages = {1--16},
  publisher = {Association for Computational Linguistics},
  address = {Bangkok, Thailand},
  doi = {10.18653/v1/2024.c3nlp-1.1},
  urldate = {2024-11-03},
}

@inproceedings{wangSeaEvalMultilingualFoundation2024,
  title = {{{SeaEval}} for Multilingual Foundation Models: From Cross-Lingual Alignment to Cultural Reasoning},
  shorttitle = {{{SeaEval}} for Multilingual Foundation Models},
  booktitle = {Proceedings of the 2024 {{Conference}} of the {{North American Chapter}} of the {{Association}} for {{Computational Linguistics}}: {{Human Language Technologies}} (Volume 1: {{Long Papers}})},
  author = {Wang, Bin and Liu, Zhengyuan and Huang, Xin and Jiao, Fangkai and Ding, Yang and Aw, AiTi and Chen, Nancy},
  editor = {Duh, Kevin and Gomez, Helena and Bethard, Steven},
  year = {2024},
  month = jun,
  pages = {370--390},
  publisher = {Association for Computational Linguistics},
  address = {Mexico City, Mexico},
  doi = {10.18653/v1/2024.naacl-long.22},
  urldate = {2025-05-01},
  langid = {english},
}

@inproceedings{qiuEvaluatingCulturalSocial2025,
  title = {Evaluating Cultural and Social Awareness of {{LLM}} Web Agents},
  booktitle = {Findings of the {{Association}} for {{Computational Linguistics}}: {{NAACL}} 2025},
  author = {Qiu, Haoyi and Fabbri, Alexander and Agarwal, Divyansh and Huang, Kung-Hsiang and Tan, Sarah and Peng, Nanyun and Wu, Chien-Sheng},
  editor = {Chiruzzo, Luis and Ritter, Alan and Wang, Lu},
  year = {2025},
  month = apr,
  pages = {3978--4005},
  publisher = {Association for Computational Linguistics},
  address = {Albuquerque, New Mexico},
  urldate = {2025-05-09},
  isbn = {979-8-89176-195-7},
  langid = {english},
}

@inproceedings{tanUnmaskingImplicitBias2025,
  title = {Unmasking Implicit Bias: Evaluating Persona-Prompted {{LLM}} Responses in Power-Disparate Social Scenarios},
  shorttitle = {Unmasking Implicit Bias},
  booktitle = {Proceedings of the 2025 {{Conference}} of the {{Nations}} of the {{Americas Chapter}} of the {{Association}} for {{Computational Linguistics}}: {{Human Language Technologies}} (Volume 1: {{Long Papers}})},
  author = {Tan, Bryan Chen Zhengyu and Lee, Roy Ka-Wei},
  editor = {Chiruzzo, Luis and Ritter, Alan and Wang, Lu},
  year = {2025},
  month = apr,
  pages = {1075--1108},
  publisher = {Association for Computational Linguistics},
  address = {Albuquerque, New Mexico},
  urldate = {2025-05-09},
  isbn = {979-8-89176-189-6},
  langid = {english},
}

@article{zhangBAVIBenchEvaluatingRobustness2025,
  title = {B-{{AVIBench}}: Toward Evaluating the Robustness of Large Vision-Language Model on Black-Box Adversarial Visual-Instructions},
  shorttitle = {B-{{AVIBench}}},
  author = {Zhang, Hao and Shao, Wenqi and Liu, Hong and Ma, Yongqiang and Luo, Ping and Qiao, Yu and Zheng, Nanning and Zhang, Kaipeng},
  year = {2025},
  month = jan,
  journal = {Trans. Info. For. Sec.},
  volume = {20},
  pages = {1434--1446},
  issn = {1556-6013},
  doi = {10.1109/TIFS.2024.3520306},
  urldate = {2025-05-09},
  langid = {english},
}

@article{leeVHELMHolisticEvaluation2024,
  title = {{{VHELM}}: A Holistic Evaluation of Vision Language Models},
  shorttitle = {Vhelm},
  author = {Lee, Tony and Tu, Haoqin and Wong, Chi H. and Zheng, Wenhao and Zhou, Yiyang and Mai, Yifan and Roberts, Josselin S. and Yasunaga, Michihiro and Yao, Huaxiu and Xie, Cihang and Liang, Percy},
  year = {2024},
  month = dec,
  journal = {Advances in Neural Information Processing Systems},
  volume = {37},
  pages = {140632--140666},
  urldate = {2025-05-09},
  langid = {english},
}

@inproceedings{nikandrouCROPEEvaluatingIncontext2025,
  title = {{{CROPE}}: Evaluating {{In-context}} Adaptation of Vision and Language Models to Culture-Specific Concepts},
  shorttitle = {Crope},
  booktitle = {Proceedings of the 2025 {{Conference}} of the {{Nations}} of the {{Americas Chapter}} of the {{Association}} for {{Computational Linguistics}}: {{Human Language Technologies}} (Volume 1: {{Long Papers}})},
  author = {Nikandrou, Malvina and Pantazopoulos, Georgios and Vitsakis, Nikolas and Konstas, Ioannis and Suglia, Alessandro},
  editor = {Chiruzzo, Luis and Ritter, Alan and Wang, Lu},
  year = {2025},
  month = apr,
  pages = {7917--7936},
  publisher = {Association for Computational Linguistics},
  address = {Albuquerque, New Mexico},
  urldate = {2025-05-09},
  isbn = {979-8-89176-189-6},
  langid = {english},
}

@inproceedings{nayakBenchmarkingVisionLanguage2024,
  title = {Benchmarking Vision Language Models for Cultural Understanding},
  booktitle = {Proceedings of the 2024 {{Conference}} on {{Empirical Methods}} in {{Natural Language Processing}}},
  author = {Nayak, Shravan and Jain, Kanishk and Awal, Rabiul and Reddy, Siva and Steenkiste, Sjoerd Van and Hendricks, Lisa Anne and Stanczak, Karolina and Agrawal, Aishwarya},
  editor = {{Al-Onaizan}, Yaser and Bansal, Mohit and Chen, Yun-Nung},
  year = {2024},
  month = nov,
  pages = {5769--5790},
  publisher = {Association for Computational Linguistics},
  address = {Miami, Florida, USA},
  doi = {10.18653/v1/2024.emnlp-main.329},
  urldate = {2025-05-09},
  langid = {english},
}

@inproceedings{kimWHENTOMEATS2025,
  title = {{{WHEN TOM EATS KIMCHI}}: Evaluating Cultural Awareness of Multimodal Large Language Models in Cultural Mixture Contexts},
  shorttitle = {When Tom Eats Kimchi},
  booktitle = {Proceedings of the 3rd {{Workshop}} on {{Cross-cultural Considerations}} in {{NLP}} ({{C3NLP}} 2025)},
  author = {Kim, Jun Seong and Thu, Kyaw Ye and Ismayilzada, Javad and Park, Junyeong and Kim, Eunsu and Ahmad, Huzama and An, Na Min and Thorne, James and Oh, Alice},
  editor = {Prabhakaran, Vinodkumar and Dev, Sunipa and Benotti, Luciana and Hershcovich, Daniel and Cao, Yong and Zhou, Li and Cabello, Laura and Adebara, Ife},
  year = {2025},
  month = may,
  pages = {143--154},
  publisher = {Association for Computational Linguistics},
  address = {Albuquerque, New Mexico},
  urldate = {2025-05-09},
  isbn = {979-8-89176-237-4},
  langid = {english},
}

@article{kannenAestheticsCulturalCompetence2024,
  title = {Beyond Aesthetics: Cultural Competence in Text-to-Image Models},
  shorttitle = {Beyond Aesthetics},
  author = {Kannen, Nithish and Ahmad, Arif and Andreetto, Marco and Prabhakaran, Vinodkumar and Prabhu, Utsav and Dieng, Adji B. and Bhattacharyya, Pushpak and Dave, Shachi},
  year = {2024},
  month = dec,
  journal = {Advances in Neural Information Processing Systems},
  volume = {37},
  pages = {13716--13747},
  urldate = {2025-05-09},
  langid = {english},
}

@inproceedings{dincaOpenBiasOpensetBias2024,
  title = {{{OpenBias}}: Open-Set Bias Detection in Text-to-Image Generative Models},
  shorttitle = {{{OpenBias}}},
  booktitle = {Proceedings of the {{IEEE}}/{{CVF Conference}} on {{Computer Vision}} and {{Pattern Recognition}}},
  author = {D'Inc{\`a}, Moreno and Peruzzo, Elia and Mancini, Massimiliano and Xu, Dejia and Goel, Vidit and Xu, Xingqian and Wang, Zhangyang and Shi, Humphrey and Sebe, Nicu},
  year = {2024},
  pages = {12225--12235},
  urldate = {2025-05-09},
  langid = {english},
}

@misc{liSurveyStateArt2025,
  title = {A Survey of State of the Art Large Vision Language Models: Alignment, Benchmark, Evaluations and Challenges},
  shorttitle = {A Survey of State of the Art Large Vision Language Models},
  author = {Li, Zongxia and Wu, Xiyang and Du, Hongyang and Liu, Fuxiao and Nghiem, Huy and Shi, Guangyao},
  year = {2025},
  month = apr,
  number = {arXiv:2501.02189},
  eprint = {2501.02189},
  primaryclass = {cs},
  publisher = {arXiv},
  doi = {10.48550/arXiv.2501.02189},
  urldate = {2025-05-09},
  archiveprefix = {arXiv},
  langid = {english},
}

@misc{HelloGPT4o2024,
  title = {Hello {{GPT-4o}}},
  author = {, OpenAI},
  year = {2024},
  month = may,
  urldate = {2024-09-28},
  howpublished = {https://openai.com/index/hello-gpt-4o/},
  langid = {american},
}

@misc{baiQwen25VLTechnicalReport2025,
  title = {Qwen2.5-{{VL}} Technical Report},
  author = {Bai, Shuai and Chen, Keqin and Liu, Xuejing and Wang, Jialin and Ge, Wenbin and Song, Sibo and Dang, Kai and Wang, Peng and Wang, Shijie and Tang, Jun and Zhong, Humen and Zhu, Yuanzhi and Yang, Mingkun and Li, Zhaohai and Wan, Jianqiang and Wang, Pengfei and Ding, Wei and Fu, Zheren and Xu, Yiheng and Ye, Jiabo and Zhang, Xi and Xie, Tianbao and Cheng, Zesen and Zhang, Hang and Yang, Zhibo and Xu, Haiyang and Lin, Junyang},
  year = {2025},
  month = feb,
  journal = {Arxiv.org},
  urldate = {2025-05-09},
  howpublished = {https://arxiv.org/abs/2502.13923v1},
  langid = {english},
}

@misc{teamKimiVLTechnicalReport2025,
  title = {Kimi-{{VL}} Technical Report},
  author = {Team, Kimi and Du, Angang and Yin, Bohong and Xing, Bowei and Qu, Bowen and Wang, Bowen and Chen, Cheng and Zhang, Chenlin and Du, Chenzhuang and Wei, Chu and Wang, Congcong and Zhang, Dehao and Du, Dikang and Wang, Dongliang and Yuan, Enming and Lu, Enzhe and Li, Fang and Sung, Flood and Wei, Guangda and Lai, Guokun and Zhu, Han and Ding, Hao and Hu, Hao and Yang, Hao and Zhang, Hao and Wu, Haoning and Yao, Haotian and Lu, Haoyu and Wang, Heng and Gao, Hongcheng and Zheng, Huabin and Li, Jiaming and Su, Jianlin and Wang, Jianzhou and Deng, Jiaqi and Qiu, Jiezhong and Xie, Jin and Wang, Jinhong and Liu, Jingyuan and Yan, Junjie and Ouyang, Kun and Chen, Liang and Sui, Lin and Yu, Longhui and Dong, Mengfan and Dong, Mengnan and Xu, Nuo and Cheng, Pengyu and Gu, Qizheng and Zhou, Runjie and Liu, Shaowei and Cao, Sihan and Yu, Tao and Song, Tianhui and Bai, Tongtong and Song, Wei and He, Weiran and Huang, Weixiao and Xu, Weixin and Yuan, Xiaokun and Yao, Xingcheng and Wu, Xingzhe and Zu, Xinxing and Zhou, Xinyu and Wang, Xinyuan and Charles, Y. and Zhong, Yan and Li, Yang and Hu, Yangyang and Chen, Yanru and Wang, Yejie and Liu, Yibo and Miao, Yibo and Qin, Yidao and Chen, Yimin and Bao, Yiping and Wang, Yiqin and Kang, Yongsheng and Liu, Yuanxin and Du, Yulun and Wu, Yuxin and Wang, Yuzhi and Yan, Yuzi and Zhou, Zaida and Li, Zhaowei and Jiang, Zhejun and Zhang, Zheng and Yang, Zhilin and Huang, Zhiqi and Huang, Zihao and Zhao, Zijia and Chen, Ziwei and Lin, Zongyu},
  year = {2025},
  month = apr,
  number = {arXiv:2504.07491},
  eprint = {2504.07491},
  primaryclass = {cs},
  publisher = {arXiv},
  doi = {10.48550/arXiv.2504.07491},
  urldate = {2025-05-09},
  archiveprefix = {arXiv},
  langid = {english},
}

@misc{leeLLaVANeXTImprovedReasoning2024,
  title = {{{LLaVA-NeXT}}: Improved Reasoning, {{OCR}}, and World Knowledge},
  shorttitle = {{{LLaVA-NeXT}}},
  author = {Liu, Haotian and Li, Chunyuan and Li, Yuheng and Li, Bo and Zhang, Yuanhan and Shen, Sheng and Lee, Yong Jae},
  year = {2024},
  month = jan,
  journal = {Llava},
  urldate = {2025-05-09},
  howpublished = {https://llava-vl.github.io/blog/2024-01-30-llava-next/},
  langid = {english},
}

@misc{Llama32Revolutionizing2024,
  title = {Llama 3.2: Revolutionizing Edge {{AI}} and Vision with Open, Customizable Models},
  shorttitle = {Llama 3.2},
  author = {, Meta AI},
  year = {2024},
  month = sep,
  journal = {Meta AI},
  urldate = {2025-05-09},
  howpublished = {https://ai.meta.com/blog/llama-3-2-connect-2024-vision-edge-mobile-devices/},
  langid = {english},
}

@misc{wuDeepSeekVL2MixtureofexpertsVisionlanguage2024,
  title = {{{DeepSeek-VL2}}: Mixture-of-Experts Vision-Language Models for Advanced Multimodal Understanding},
  shorttitle = {{{DeepSeek-VL2}}},
  author = {Wu, Zhiyu and Chen, Xiaokang and Pan, Zizheng and Liu, Xingchao and Liu, Wen and Dai, Damai and Gao, Huazuo and Ma, Yiyang and Wu, Chengyue and Wang, Bingxuan and Xie, Zhenda and Wu, Yu and Hu, Kai and Wang, Jiawei and Sun, Yaofeng and Li, Yukun and Piao, Yishi and Guan, Kang and Liu, Aixin and Xie, Xin and You, Yuxiang and Dong, Kai and Yu, Xingkai and Zhang, Haowei and Zhao, Liang and Wang, Yisong and Ruan, Chong},
  year = {2024},
  month = dec,
  number = {arXiv:2412.10302},
  eprint = {2412.10302},
  primaryclass = {cs},
  publisher = {arXiv},
  doi = {10.48550/arXiv.2412.10302},
  urldate = {2025-05-09},
  archiveprefix = {arXiv},
  langid = {english},
}

@misc{steinerPaliGemma2Family2024,
  title = {{{PaliGemma}} 2: A Family of Versatile {{VLMs}} for Transfer},
  shorttitle = {{{PaliGemma}} 2},
  author = {Steiner, Andreas and Pinto, Andr{\'e} Susano and Tschannen, Michael and Keysers, Daniel and Wang, Xiao and Bitton, Yonatan and Gritsenko, Alexey and Minderer, Matthias and Sherbondy, Anthony and Long, Shangbang and Qin, Siyang and Ingle, Reeve and Bugliarello, Emanuele and Kazemzadeh, Sahar and Mesnard, Thomas and Alabdulmohsin, Ibrahim and Beyer, Lucas and Zhai, Xiaohua},
  year = {2024},
  month = dec,
  number = {arXiv:2412.03555},
  eprint = {2412.03555},
  primaryclass = {cs},
  publisher = {arXiv},
  doi = {10.48550/arXiv.2412.03555},
  urldate = {2025-05-09},
  archiveprefix = {arXiv},
  langid = {english},
}

@misc{zhuInternVL3ExploringAdvanced2025,
  title = {{{InternVL3}}: Exploring Advanced Training and Test-Time Recipes for Open-Source Multimodal Models},
  shorttitle = {{{InternVL3}}},
  author = {Zhu, Jinguo and Wang, Weiyun and Chen, Zhe and Liu, Zhaoyang and Ye, Shenglong and Gu, Lixin and Tian, Hao and Duan, Yuchen and Su, Weijie and Shao, Jie and Gao, Zhangwei and Cui, Erfei and Wang, Xuehui and Cao, Yue and Liu, Yangzhou and Wei, Xingguang and Zhang, Hongjie and Wang, Haomin and Xu, Weiye and Li, Hao and Wang, Jiahao and Deng, Nianchen and Li, Songze and He, Yinan and Jiang, Tan and Luo, Jiapeng and Wang, Yi and He, Conghui and Shi, Botian and Zhang, Xingcheng and Shao, Wenqi and He, Junjun and Xiong, Yingtong and Qu, Wenwen and Sun, Peng and Jiao, Penglong and Lv, Han and Wu, Lijun and Zhang, Kaipeng and Deng, Huipeng and Ge, Jiaye and Chen, Kai and Wang, Limin and Dou, Min and Lu, Lewei and Zhu, Xizhou and Lu, Tong and Lin, Dahua and Qiao, Yu and Dai, Jifeng and Wang, Wenhai},
  year = {2025},
  month = apr,
  number = {arXiv:2504.10479},
  eprint = {2504.10479},
  primaryclass = {cs},
  publisher = {arXiv},
  doi = {10.48550/arXiv.2504.10479},
  urldate = {2025-05-09},
  archiveprefix = {arXiv},
  langid = {english},
}

@misc{deitkeMolmoPixMoOpen2024,
  title = {Molmo and {{PixMo}}: Open Weights and Open Data for State-of-the-Art Vision-Language Models},
  shorttitle = {Molmo and {{PixMo}}},
  author = {Deitke, Matt and Clark, Christopher and Lee, Sangho and Tripathi, Rohun and Yang, Yue and Park, Jae Sung and Salehi, Mohammadreza and Muennighoff, Niklas and Lo, Kyle and Soldaini, Luca and Lu, Jiasen and Anderson, Taira and Bransom, Erin and Ehsani, Kiana and Ngo, Huong and Chen, YenSung and Patel, Ajay and Yatskar, Mark and {Callison-Burch}, Chris and Head, Andrew and Hendrix, Rose and Bastani, Favyen and VanderBilt, Eli and Lambert, Nathan and Chou, Yvonne and Chheda, Arnavi and Sparks, Jenna and Skjonsberg, Sam and Schmitz, Michael and Sarnat, Aaron and Bischoff, Byron and Walsh, Pete and Newell, Chris and Wolters, Piper and Gupta, Tanmay and Zeng, Kuo-Hao and Borchardt, Jon and Groeneveld, Dirk and Nam, Crystal and Lebrecht, Sophie and Wittlif, Caitlin and Schoenick, Carissa and Michel, Oscar and Krishna, Ranjay and Weihs, Luca and Smith, Noah A. and Hajishirzi, Hannaneh and Girshick, Ross and Farhadi, Ali and Kembhavi, Aniruddha},
  year = {2024},
  month = dec,
  number = {arXiv:2409.17146},
  eprint = {2409.17146},
  primaryclass = {cs},
  publisher = {arXiv},
  doi = {10.48550/arXiv.2409.17146},
  urldate = {2025-05-09},
  archiveprefix = {arXiv},
  langid = {english},
}

@misc{liuNVILAEfficientFrontier2025,
  title = {{{NVILA}}: Efficient Frontier Visual Language Models},
  shorttitle = {Nvila},
  author = {Liu, Zhijian and Zhu, Ligeng and Shi, Baifeng and Zhang, Zhuoyang and Lou, Yuming and Yang, Shang and Xi, Haocheng and Cao, Shiyi and Gu, Yuxian and Li, Dacheng and Li, Xiuyu and Fang, Yunhao and Chen, Yukang and Hsieh, Cheng-Yu and Huang, De-An and Cheng, An-Chieh and Nath, Vishwesh and Hu, Jinyi and Liu, Sifei and Krishna, Ranjay and Xu, Daguang and Wang, Xiaolong and Molchanov, Pavlo and Kautz, Jan and Yin, Hongxu and Han, Song and Lu, Yao},
  year = {2025},
  month = mar,
  number = {arXiv:2412.04468},
  eprint = {2412.04468},
  primaryclass = {cs},
  publisher = {arXiv},
  doi = {10.48550/arXiv.2412.04468},
  urldate = {2025-05-09},
  archiveprefix = {arXiv},
  langid = {english},
}

@inproceedings{huLoRALowrankAdaptation2021,
  title = {{{LoRA}}: Low-Rank Adaptation of Large Language Models},
  shorttitle = {{{LoRA}}},
  booktitle = {International {{Conference}} on {{Learning Representations}}},
  author = {Hu, Edward J. and Shen, Yelong and Wallis, Phillip and {Allen-Zhu}, Zeyuan and Li, Yuanzhi and Wang, Shean and Wang, Lu and Chen, Weizhu},
  year = {2021},
  month = oct,
  urldate = {2025-05-12},
  langid = {english},
}

@misc{satarSeeingCultureBenchmark2025,
  title = {Seeing Culture: A Benchmark for Visual Reasoning and Grounding},
  shorttitle = {Seeing Culture},
  author = {Satar, Burak and Ma, Zhixin and Irawan, Patrick A. and Mulyawan, Wilfried A. and Jiang, Jing and Lim, Ee-Peng and Ngo, Chong-Wah},
  year = {2025},
  month = sep,
  number = {arXiv:2509.16517},
  eprint = {2509.16517},
  primaryclass = {cs},
  publisher = {arXiv},
  doi = {10.48550/arXiv.2509.16517},
  urldate = {2025-09-30},
  archiveprefix = {arXiv},
  langid = {english},
}

@misc{fortinIntroducingGemini252025,
  title = {Introducing Gemini 2.5 Flash Image, Our State-of-the-Art Image Model- Google Developers Blog},
  author = {Fortin, Alias and Vernade, Guillaume and Kampf, Kat and Reshi, Ammaar},
  year = {2025},
  month = aug,
  urldate = {2025-09-30},
  howpublished = {https://developers.googleblog.com/en/introducing-gemini-2-5-flash-image/},
  langid = {english}
}

@misc{winataWorldCuisinesMassivescaleBenchmark2025,
  title = {{{WorldCuisines}}: A Massive-Scale Benchmark for Multilingual and Multicultural Visual Question Answering on Global Cuisines},
  shorttitle = {{{WorldCuisines}}},
  author = {Winata, Genta Indra and Hudi, Frederikus and Irawan, Patrick Amadeus and Anugraha, David and Putri, Rifki Afina and Wang, Yutong and Nohejl, Adam and Prathama, Ubaidillah Ariq and Ousidhoum, Nedjma and Amriani, Afifa and Rzayev, Anar and Das, Anirban and Pramodya, Ashmari and Adila, Aulia and Wilie, Bryan and Mawalim, Candy Olivia and Cheng, Ching Lam and Abolade, Daud and Chersoni, Emmanuele and Santus, Enrico and Ikhwantri, Fariz and Kuwanto, Garry and Zhao, Hanyang and Wibowo, Haryo Akbarianto and Lovenia, Holy and Cruz, Jan Christian Blaise and Putra, Jan Wira Gotama and Myung, Junho and Susanto, Lucky and Machin, Maria Angelica Riera and Zhukova, Marina and Anugraha, Michael and Adilazuarda, Muhammad Farid and Santosa, Natasha and Limkonchotiwat, Peerat and Dabre, Raj and Audino, Rio Alexander and Cahyawijaya, Samuel and Zhang, Shi-Xiong and Salim, Stephanie Yulia and Zhou, Yi and Gui, Yinxuan and Adelani, David Ifeoluwa and Lee, En-Shiun Annie and Okada, Shogo and Purwarianti, Ayu and Aji, Alham Fikri and Watanabe, Taro and Wijaya, Derry Tanti and Oh, Alice and Ngo, Chong-Wah},
  year = {2025},
  month = may,
  number = {arXiv:2410.12705},
  eprint = {2410.12705},
  primaryclass = {cs},
  publisher = {arXiv},
  doi = {10.48550/arXiv.2410.12705},
  urldate = {2025-10-04},
  archiveprefix = {arXiv},
  langid = {english}
}

@inproceedings{chiuCulturalBenchRobustDiverse2025,
  title = {{{CulturalBench}}: A Robust, Diverse and Challenging Benchmark for Measuring {{LMs}}' Cultural Knowledge through Human-{{AI}} Red-Teaming},
  shorttitle = {{{CulturalBench}}},
  booktitle = {Proceedings of the 63rd {{Annual Meeting}} of the {{Association}} for {{Computational Linguistics}} ({{Volume}} 1: {{Long Papers}})},
  author = {Chiu, Yu Ying and Jiang, Liwei and Lin, Bill Yuchen and Park, Chan Young and Li, Shuyue Stella and Ravi, Sahithya and Bhatia, Mehar and Antoniak, Maria and Tsvetkov, Yulia and Shwartz, Vered and Choi, Yejin},
  editor = {Che, Wanxiang and Nabende, Joyce and Shutova, Ekaterina and Pilehvar, Mohammad Taher},
  year = 2025,
  month = jul,
  pages = {25663--25701},
  publisher = {Association for Computational Linguistics},
  address = {Vienna, Austria},
  doi = {10.18653/v1/2025.acl-long.1247},
  urldate = {2026-01-24},
  isbn = {979-8-89176-251-0},
  langid = {english},
}

@misc{romeroCVQACulturallydiverseMultilingual2024,
  title = {{{CVQA}}: Culturally-Diverse Multilingual Visual Question Answering Benchmark},
  shorttitle = {Cvqa},
  author = {Romero, David and Lyu, Chenyang and Wibowo, Haryo Akbarianto and Lynn, Teresa and Hamed, Injy and Kishore, Aditya Nanda and Mandal, Aishik and Dragonetti, Alina and Abzaliev, Artem and Tonja, Atnafu Lambebo and Balcha, Bontu Fufa and Whitehouse, Chenxi and Salamea, Christian and Velasco, Dan John and Adelani, David Ifeoluwa and Meur, David Le and {Villa-Cueva}, Emilio and Koto, Fajri and Farooqui, Fauzan and Belcavello, Frederico and Batnasan, Ganzorig and Vallejo, Gisela and Caulfield, Grainne and Ivetta, Guido and Song, Haiyue and Ademtew, Henok Biadglign and Maina, Hern{\'a}n and Lovenia, Holy and Azime, Israel Abebe and Cruz, Jan Christian Blaise and Gala, Jay and Geng, Jiahui and {Ortiz-Barajas}, Jesus-German and Baek, Jinheon and Dunstan, Jocelyn and Alemany, Laura Alonso and Nagasinghe, Kumaranage Ravindu Yasas and Benotti, Luciana and D'Haro, Luis Fernando and Viridiano, Marcelo and {Estecha-Garitagoitia}, Marcos and Cabrera, Maria Camila Buitrago and {Rodr{\'i}guez-Cantelar}, Mario and Jouitteau, M{\'e}lanie and Mihaylov, Mihail and Imam, Mohamed Fazli Mohamed and Adilazuarda, Muhammad Farid and Gochoo, Munkhjargal and Otgonbold, Munkh-Erdene and Etori, Naome and Niyomugisha, Olivier and Silva, Paula M{\'o}nica and Chitale, Pranjal and Dabre, Raj and Chevi, Rendi and Zhang, Ruochen and Diandaru, Ryandito and Cahyawijaya, Samuel and G{\'o}ngora, Santiago and Jeong, Soyeong and Purkayastha, Sukannya and Kuribayashi, Tatsuki and Clifford, Teresa and Jayakumar, Thanmay and Torrent, Tiago Timponi and Ehsan, Toqeer and Araujo, Vladimir and Kementchedjhieva, Yova and Burzo, Zara and Lim, Zheng Wei and Yong, Zheng Xin and Ignat, Oana and Nwatu, Joan and Mihalcea, Rada and Solorio, Thamar and Aji, Alham Fikri},
  year = {2024},
  month = nov,
  number = {arXiv:2406.05967},
  eprint = {2406.05967},
  primaryclass = {cs},
  publisher = {arXiv},
  doi = {10.48550/arXiv.2406.05967},
  urldate = {2025-10-05},
  archiveprefix = {arXiv},
  langid = {english}
}

@misc{liRAVENEABenchmarkMultimodal2025,
  title = {{{RAVENEA}}: A Benchmark for Multimodal Retrieval-Augmented Visual Culture Understanding},
  shorttitle = {Ravenea},
  author = {Li, Jiaang and Yuan, Yifei and Li, Wenyan and Aliannejadi, Mohammad and Hershcovich, Daniel and S{\o}gaard, Anders and Vuli{\'c}, Ivan and Zhang, Wenxuan and Liang, Paul Pu and Deng, Yang and Belongie, Serge},
  year = 2025,
  month = may,
  number = {arXiv:2505.14462},
  eprint = {2505.14462},
  primaryclass = {cs},
  publisher = {arXiv},
  doi = {10.48550/arXiv.2505.14462},
  urldate = {2026-01-13},
  archiveprefix = {arXiv},
  langid = {english},
}

@inproceedings{alwajihPalmCulturallyInclusive2025,
  title = {Palm: A Culturally Inclusive and Linguistically Diverse Dataset for Arabic {{LLMs}}},
  shorttitle = {Palm},
  booktitle = {Proceedings of the 63rd {{Annual Meeting}} of the {{Association}} for {{Computational Linguistics}} ({{Volume}} 1: {{Long Papers}})},
  author = {Alwajih, Fakhraddin and El Mekki, Abdellah and Magdy, Samar Mohamed and Elmadany, AbdelRahim A. and Nacar, Omer and Nagoudi, El Moatez Billah and {Abdel-Salam}, Reem and Atwany, Hanin and Nafea, Youssef and Yahya, Abdulfattah Mohammed and Alhamouri, Rahaf and Alsayadi, Hamzah A. and Zayed, Hiba and Shatnawi, Sara and Sibaee, Serry and {Ech-chammakhy}, Yasir and {Al-Dhabyani}, Walid and Ali, Marwa Mohamed and Jarraya, Imen and {El-Shangiti}, Ahmed Oumar and Alraeesi, Aisha and {AL-Ghrawi}, Mohammed Anwar and {Al-Batati}, Abdulrahman S. and Mohamed, Elgizouli and Elgindi, Noha Taha and Saeed, Muhammed and Atou, Houdaifa and Yahia, Issam Ait and Bouayad, Abdelhak and Machrouh, Mohammed and Makouar, Amal and Alkawi, Dania and Mohamed, Mukhtar and Abdelfadil, Safaa Taher and Ounnoughene, Amine Ziad and Rouabhia, Anfel and Assi, Rwaa and Sorkatti, Ahmed and Tourad, Mohamedou Cheikh and Koubaa, Anis and Berrada, Ismail and Jarrar, Mustafa and Shehata, Shady and {Abdul-Mageed}, Muhammad},
  editor = {Che, Wanxiang and Nabende, Joyce and Shutova, Ekaterina and Pilehvar, Mohammad Taher},
  year = 2025,
  month = jul,
  pages = {32871--32894},
  publisher = {Association for Computational Linguistics},
  address = {Vienna, Austria},
  doi = {10.18653/v1/2025.acl-long.1579},
  urldate = {2026-01-13},
  isbn = {979-8-89176-251-0},
  langid = {english},
}
